%% file: arxiv_version.tex
\documentclass[journal]{IEEEtran}
\ifCLASSOPTIONcompsoc
  \usepackage[nocompress]{cite}
\else
  \usepackage{cite}
\fi
\ifCLASSINFOpdf
\else
\fi
\usepackage{amsmath}
\usepackage{amssymb}
\usepackage{subfigure}
\usepackage{caption}
\usepackage{multirow}
\usepackage{array}
\usepackage{algorithm}
\usepackage{algorithmic}
\usepackage[percent]{overpic}
\usepackage{color}
\usepackage{soul}

\graphicspath{{./figures/Latex_pictures/}{./figures/GAN_results/}{./figures/generated/}{./figures/PG_GAN_results/}{./figures/sample_blurs/}}
\include{Ahmed-defs}

\begin{document}
%
\title{Blind Image Deconvolution using Deep Generative Priors}
%
%

%
%

\author{Muhammad~Asim\textsuperscript{*} ,
        Fahad~Shamshad\textsuperscript{*},
        and~Ali~Ahmed \thanks{Muhammad Asim, Fahad Shamshad, and Ali Ahmed are with Department of Electrical Engineering, Information Technology University, Lahore, Pakistan, email: \{msee16001, fahad.shamshad, ali.ahmed\}@itu.edu.pk.} \thanks{* Authors contributed equally to this work.} }
\IEEEtitleabstractindextext{%
\begin{abstract}
This paper proposes a novel approach to regularize the \textit{ill-posed} and \textit{non-linear} blind image deconvolution (blind deblurring) using deep generative networks as priors. We employ two separate generative models --- one trained to produce sharp images while the other trained to generate blur kernels from lower-dimensional parameters.  To deblur, we propose an alternating gradient descent scheme operating in the latent lower-dimensional space of each of the pretrained generative models. Our experiments show promising deblurring results on  images even under large blurs, and heavy noise. 
To address the shortcomings of generative models such as mode collapse, we augment our generative priors with classical image priors and report improved performance on complex image datasets. The deblurring performance depends on how well the range of the generator spans the image class. Interestingly, our experiments show that even an untrained structured (convolutional) generative networks acts as an image prior in the image deblurring context allowing us to extend our results to more diverse natural image datasets. 
\end{abstract}

\begin{IEEEkeywords}
Blind image deblurring, generative adversarial networks, variational autoencoders, deep image prior.
\end{IEEEkeywords}}

\maketitle

\IEEEdisplaynontitleabstractindextext
%
\IEEEpeerreviewmaketitle

\section{Introduction}\label{sec:introduction}

\IEEEPARstart{B}{lind} image deblurring aims to recover a true image $i$ and a blur kernel $k$ from blurry and possibly noisy observation $y$. For a uniform and spatially invariant blur, it can be mathematically formulated as 
\begin{equation} \label{eq:bid}
y = i \otimes k + n,
\end{equation}
where $\otimes $ is a convolution operator and $n$ is an additive Gaussian noise. In its full generality, the inverse problem \eqref{eq:bid}  is severely ill-posed as many different instances of $i$, and $k$ fit the observation $y$; see, \cite{campisi2016blind,kundur1996blind} for a thorough discussion on solution ambiguities in blind deconvolution. 

To resolve between multiple instances, priors are introduced on images and/or blur kernels in the image deblurring algorithms. Priors assume an \textit{a priori} model on the true image/blur kernel or both. Conventional priors include sparsity of the true image and/or blur kernel in some transform domain such as wavelets, curvelets, etc,  sparsity of image gradients \cite{chan1998total,fergus2006removing}, $\ell_0$ regularized prior \cite{xu2013unnatural}, internal patch recurrence \cite{michaeli2014blind}, low-rank \cite{ahmed2014blind,ren2016image}, and hyperlaplacian prior \cite{krishnan2009fast}, etc. Although generic and applicable to multiple applications, these engineered models are not very effective as many unrealistic images also fit the prior model.

\begin{figure}[t]
\centering
\includegraphics[width=\columnwidth]{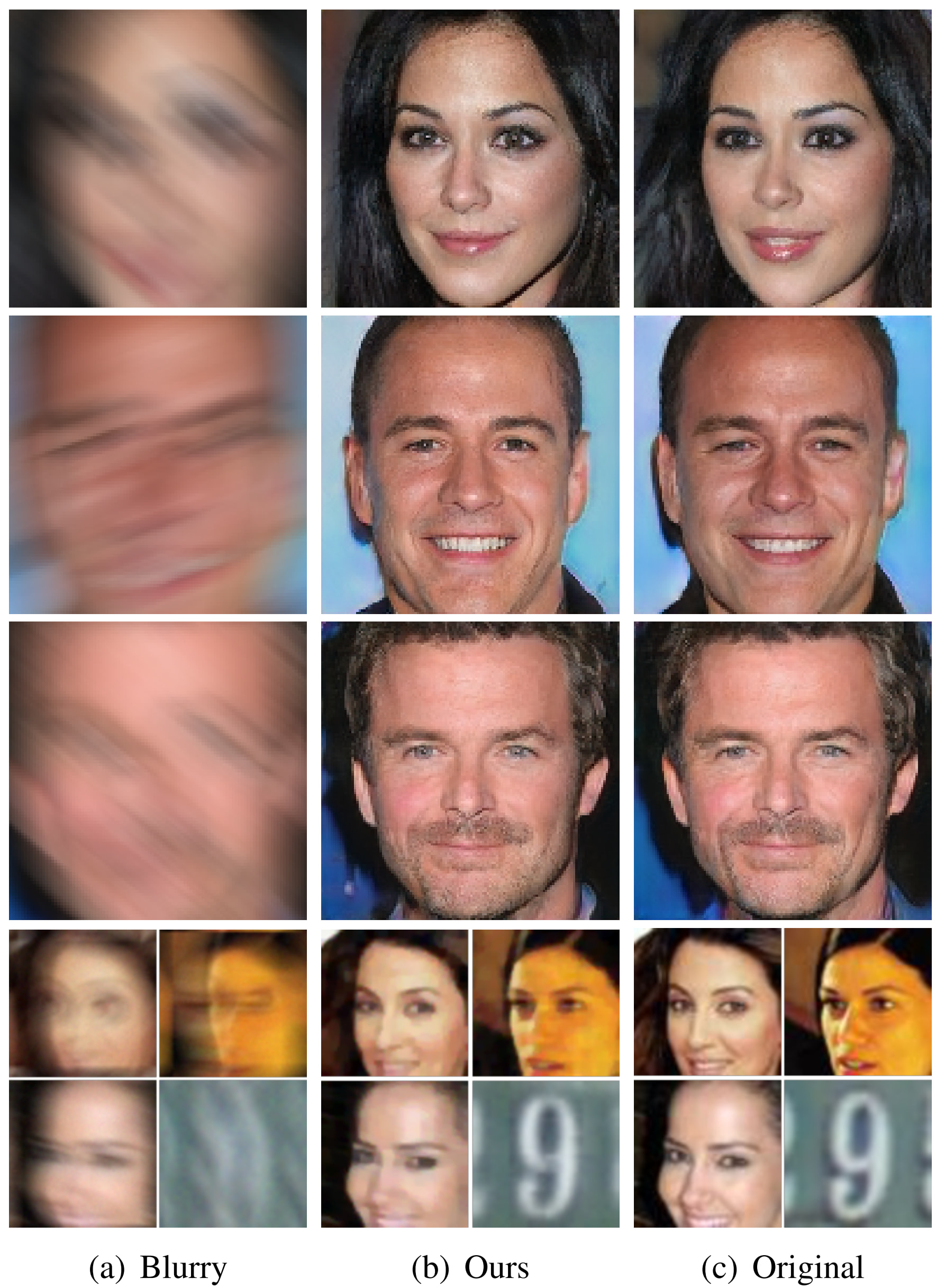}
\caption{ \small  Blind image deblurring using deep generative priors.}
\label{fig:intro-results}
\end{figure}

Recently, deep learning has emerged as a new state of the art in blind image deconvolution like in many other image restoration problems. The results so far focus on bypassing the blur kernel estimation, and training a network in an end-to-end manner to fit blurred images with corresponding sharp ones \cite{schuler2016learning,hradivs2015convolutional,chakrabarti2016neural,svoboda2016cnn}. The main drawback of this end-to-end deep learning approach is that it does not explicitly take into account the knowledge of forward map or the governing equation \eqref{eq:bid}, but rather learns implicitly from training data. Consequently, the deblurring is more sensitive to changes in the blur kernels, images, or noise distributions in the test set that are not representative of the training data, and often requires expensive retraining of the network for a competitive performance. Comparatively, this paper seeks to deblur images by employing generative networks in a novel role of priors in the inverse problem.

In last couple of years, advances in implicit generative modeling of images \cite{hand2017global} have taken it well beyond the conventional prior models outlined above. The introduction of such deep generative priors in the image deblurring should enable a far more effective regularization yielding sharper and better quality deblurred images. Our experiments in Figure \ref{fig:intro-results} confirm this hypothesis, and show that embedding pretrained deep generative priors in an iterative scheme for image deblurring produces promising results on standard datasets of face images and house numbers. Some of the blurred faces are almost not recognizable by the naked eye due to extravagant blurs yet the recovery algorithm deblurs these images near perfectly with the assistance of generative priors.

This paper shows that an alternating gradient decent scheme assisted with generative priors is able to recover the blur kernel $k$, and a visually appealing and sharp approximation of the true image $i$ from the blurry $y$. Specifically, the algorithm searches for a pair  $(\hat{i}, \hat{k})$ in the range of respective pretrained generators of images and blur kernels that explains the blurred image $y$. Implicitly, the generative priors aggressively regularize the alternating gradient descent algorithm to produce a sharp and a clean image. Since the range of the generative models can be traversed by a much lower dimensional latent representations compared to the ambient dimension of the images, it not only reduces the number of unknowns in the deblurring problem but also allows for an efficient implementation of gradients in this lower dimensional space using back propagation through the generators. 

The numerical experiments manifest that, in general, the deep generative priors yield better deblurring results compared to the conventional image priors, studied extensively in the literature. Compared to end-to-end deep learning based frameworks, our approach explicitly takes into account the knowledge of forward map (convolution) in the gradient decent algorithm to achieve robust results. Moreover, our approach does not require expensive retraining of every deep network involved in case of partial changes in blur problem specifications such as alterations in the blur kernels or noise models; in the former case, we only have to retrain the blur generative model (a shallow network, and hence easy to train), and no change in the later case.

We have found that often strictly constraining the recovered image to lie in the generator range might be counter productive owing to the limitation of the generative models to faithfully learn the image distribution: reasons may include mode collapse and convergence issues among others. We, therefore, investigate a modification of the loss function to allow the recovered images some leverage/slack to deviate from the range of the generator. This modification effectively addresses the performance limitation due to the range of the generator. 

Another important contribution of this work is to show that even untrained deep convolutional generative networks can act as a good image prior in blind image deblurring. This learning free prior ability suggests that some of the image statistics are captured by the network structure alone. The weights of the untrained network are initialized using a single blurred image. The fact that untrained generative models can act as good priors \cite{ulyanov2017deep}, allows us to importantly elevate the performance of our algorithm on rich image datasets that are currently not effectively learned by the generative models.

The rest of the paper is organized as follows. In Section \ref{relatedwork}, we give an overview of the related work. We formulate the problem in Section \ref{sec:Problem-Formulation} followed by our proposed alternating gradient descent algorithms in Section \ref{sec:proposedapp}. Section \ref{experiments} contains experimental results followed by concluding remarks in Section \ref{sec:conc}.
 

\section{Related Work}
\label{relatedwork}

Blind image deblurring is a well-studied topic and in general, various priors/regularizers exploiting the structure of an image or blur kernel are introduced to address the ill-posedness. These natural structures expect images or blur kernels to be sparse in some transform domain; see, for example, \cite{chan1998total, fergus2006removing, levin2009understanding,hu2010single, zhang2011sparse,cai2009blind}.  Another approach \cite{hu2010single, zhang2011sparse} is to learn an over complete dictionary for sparse representations of image patches.  The inverse problem is regularized to favor sparsely representable image patches in the learned dictionary. On the other hand, we learn a more powerful non-linear mapping (generative network) of full size images to lower-dimensional feature vectors. The inverse problem is now regularized by constraining the images in the range of the generator. Some of the other penalty functions to improve the conditioning of the blind image deblurring problem are low-rank \cite{ren2016image} and total variation \cite{pan2014motion} based priors. A recently introduced dark channel prior \cite{pan2016blind} also shows promising results; it assumes a sparse structure on the dark channel of the image, and exploits this structure in an optimization program \cite{xu2011image} for the inverse problem. Other works include extreme channel priors \cite{yan2017image}, outlier robust deblurring \cite{dong2017blind}, learned data fitting \cite{pan2017learning}, and discriminative prior based blind image deblurring approaches \cite{li2018learning}.

Our approach bridges the gap between the conventional iterative schemes, and  recent end-to-end deep networks for image deblurring \cite{hradivs2015convolutional,nah2016deep,schuler2016learning,xu2017learning,nimisha2017blur,kupyn2017deblurgan, chen2018motion}. The iterative schemes are generally adaptable to new images/blurs, and other modifications in the model such as noise statistics. Comparatively, the end-to-end approaches above breakdown to any such changes that are not reflected in the training data, and require a complete retraining of the network. Our approach combines some of the benefits of both these paradigms by employing powerful generative neural networks in an iterative scheme as priors that already are \textit{familiar} with images and blurs under consideration. These neural networks help restrict the candidate solutions to come from the learned or \textit{familiar} images, and blur kernels only. A change in, for example,  blur kernel model only requires retraining of a shallow network of blurs while keeping the image generative network, and rest of the iterative scheme unchanged. Similarly, a change in noise statistics is handled in a complete adaptable manner as in classical iterative schemes. Retaining the adaptability while maintaining a strong influence of the powerful neural networks is an important feature of our algorithm.

Neural network based implicit generative models such as generative adversarial networks (GANs) \cite{goodfellow2014generative} and variational autoencoders (VAEs) \cite{kingma2013auto}  have found much success in modeling complex data distributions especially that of images. Recently, GANs and VAEs have been used for blind image deblurring but only in an end-to end manner, which is completely different from our approach as discussed above in detail. In \cite{xu2017learning}, authors jointly deblurs and super-resolves low resolution blurry text and face images by introducing a novel feature matching loss term during training process of GAN. In \cite{nimisha2017blur}, authors proposed sparse coding based framework consisting of both variational learning, that learns data prior by encoding its features into compact form, and adversarial learning for discriminating clean and blurry image features. A conditional GAN has been employed by \cite{kupyn2017deblurgan}  for blind motion deblurring in an end to end framework and optimized it using a multi-component loss consisting of both content and adversarial terms. These methods show competitive performance, but since these generative model based approaches are end-to-end they suffer from the same draw backs as other deep learning techniques; discussed in detail above.

The generative priors have recently been employed in solving inverse problems such as compressed sensing \cite{bora2017compressed,shah2018solving}, phase retrieval \cite{hand2018phase, shamshad2018robust}, Fourier ptychography \cite{shamshad2018deep}, and image inpainting \cite{yeh2017semantic}, etc. We also note work of \cite{samangouei2018defense} and \cite{ilyas2017robust} that use pretrained generators for circumventing the issue of adversarial attacks. To the best of our knowledge, our work is the first instance of using pretrained generative models as priors for solving blind image deconvolution .

\section{Problem Formulation}\label{sec:Problem-Formulation}

We assume the image $i \in \R^n$ and blur kernel $k \in \R^n$ in \eqref{eq:bid} are members of some structured classes $\mathcal{I}$ of images, and $\mathcal{K}$ of blurs, respectively. For example, $\setI$ may be a set of celebrity faces and $\setK$ comprises of  motion blurs. A representative sample set from both classes $\setI$ and $\setK$ is employed to train a generative model for each class. We denote by the mappings $G_{\mathcal{I}}: \mathbb{R}^l \rightarrow \mathbb{R}^n$ and $G_{\mathcal{K}}: \mathbb{R}^m \rightarrow \mathbb{R}^n$, the generators for class $\setI$, and $\setK$, respectively. Given low-dimensional inputs $z_i \in \mathbb{R}^l$, and $z_k \in \mathbb{R}^m$, the pretrained generators $G_{\setI}$, and $G_{\setK}$ generate new samples $G_{\setI}(z_i)$, and $G_{\setK}(z_k)$ that are representative of the classes $\setI$, and $\setK$, respectively. Once trained, the weights of the generators are fixed. To recover the sharp image, and blur kernel $(i,k)$ from the blurred image $y$ in \eqref{eq:bid}, we propose minimizing the following objective function 
\begin{align}\label{eq:Optimization-Ambient}
(\hat{i},\hat{k}) :=  \underset{\substack{i \in\text{Range}(G_{\setI}) \\k \in\text{Range}(G_{\setK})}}{\text{argmin}} \ \|y - i \otimes k \|^2, 
\end{align}
where Range($G_\setI$) and Range($G_\setK$) is the set of all the images and blurs that can be generated by $G_\setI$ and $G_\setK$, respectively. In words, we want to find an image $i$ and a blur kernel $k$ in the range of their respective generators, that best explain the forward model \eqref{eq:bid}. Ideally, the range of a pretrained generator comprises of only the samples drawn from the distribution of the image or blur class. Constraining the solution $(\hat{i},\hat{k})$ to lie only in generator ranges, therefore, implicitly reduces the solution ambiguities inherent to the ill-posed blind deconvolution problem, and forces the solution to be the members of classes $\setI$, and $\setK$.

The minimization program in \eqref{eq:Optimization-Ambient} can be equivalently formulated in the lower dimensional, latent representation space as follows
\begin{align}\label{eq:Optimization-latent}
(\hat{z}_i, \hat{z}_k) = \underset{z_i \in \R^l, z_k \in \R^m}{\text{argmin}}
\ 
\| y - G_{\mathcal{I}}(z_i) \otimes G_\setK(z_k) \|^2.
\end{align}
This optimization program can be thought of as tweaking the latent representation vectors $z_i$, and $z_k$, (input to the generators $G_{\setI}$, and $G_{\setK}$, respectively)  until these generators generate an image $i$ and blur kernel $k$ whose convolution comes as close to $y$ as possible. 

The optimization program in \eqref{eq:Optimization-latent} is obviously non-convex owing to the bilinear convolution operator, and non-linear deep generative models. We resort to an alternating gradient descent algorithm to find a local minima  $(\hat{z}_i, \hat{z}_k)$.  Importantly, the weights of the generators are always fixed as they enter into this algorithm as pretrained models.  At a given iteration, we fix $z_i$ and take a descent step in $z_k$, and vice verse. The gradient step in each variable involves a forward and backward pass through the generator networks. Section \ref{proposedapproach} talks about the back propagation, and gives explicit gradient forms for descent in each $z_i$ and $z_k$ for this particular algorithm. 

The estimated deblurred image and the blur kernel are acquired by a forward pass of the solutions $\hat{z}_i$ and $\hat{z}_k$ through the generators $G_{\mathcal{I}}$ and $G_{\mathcal{K}}$. Mathematically, $(\hat{i},\hat{k}) = ( G_\mathcal{I}(\hat{z}_i), G_\mathcal{K}(\hat{z}_k))$.

\section{Image Deblurring Algorithm} \label{sec:proposedapp}
Our approach requires pretrained generative models $G_\setI$ and $G_\setK$ for classes $\setI$ and $\setK$, respectively. We use both GANs and VAEs as generative models on the clean images and blur kernels. We briefly recap the training process GANs and VAEs below.

\begin{figure*}[t]
\centering
\includegraphics[width=0.90\textwidth]{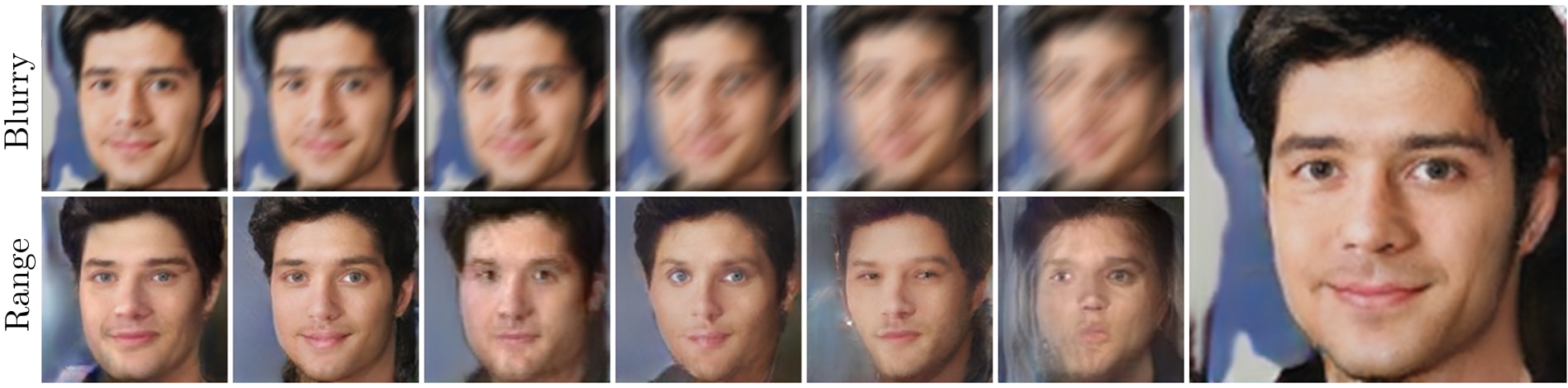}
\caption{ \small Naive Deblurring. We gradually increase blur size from left to right and demonstrate the failure of naive deblurring by finding the closest image in the range of the image generator (last row) to blurred image.  }
\label{fig:Back-Propagation}
\end{figure*}

\subsection{Training the Generative Models} \label{generative}

A generative model\footnote{The discussion in this section applies to both $G_{\setI}$, and $G_{\setK}$, therefore, we ignore the subscripts.} $G$ will either be trained via adversarial learning \cite{goodfellow2014generative} or variational inference \cite{kingma2013auto}.

Generative adversarial networks (GANs) learn the distribution $p(i)$ of images in class $\setI$ by playing an adversarial game. A discriminator network $D$ learns to differentiate between true images sampled from $p(i)$ and fake images of the generator network $G$, while $G$ tries to fool the discriminator. The cost function describing the game is given by 
\begin{align*}
\underset{G}{\text{min}} \, \underset{D}{\text{max}} \ \E_{p(i)} \log D (i) + \E_{p(z)}\log(1 - D(G(z))),
\end{align*}  
where $p(z)$ is the distribution of latent random variables $z$, and is usually defined to be a known and simple distribution such as $p(z) = \setN(0,I)$. 

Variational autoencoders (VAE) learn the distribution $p(i)$ by maximizing a lower bound on the log likelihood: 
\begin{align*}
\log p(i) \geq \E_{q(z|i)} \log p(i | z) - \text{KL}(q(z| i)\| p(z)),
\end{align*}
where the second term on the right hand side is the Kullback-Leibler divergence between known distribution $p(z)$, and  $q(z|i)$. The distribution $q(z|i)$ is a proxy for the unknown $p(z|i)$.  Under a rich enough function model $q(z|i)$, the lower bound is expected to be tight. The functional forms of $q(z|i)$, and $p(i|z)$ each are modeled via a deep network. The right hand side is maximized by tweaking the network parameters. The deep network $p(i|z)$ is the generative model that produces samples of $p(i)$ from latent representation $z$. 

Generative model $G_{\set{I}}$ for the face and shoe images is trained using adversarial learning for visually better quality results. Each of the generative model $G_{\setI}$ for other image datasets, and $G_{\setK}$ for blur kernels are trained using variational inference framework above.

\subsection{Naive Deblurring} \label{backprop}
To deblur an image $y$, a simplest possible strategy is to find an image closest to $y$ in the range of the given generator $G_{\setI}$ of clean images. Mathematically, this amounts to solving the following optimization program 
\begin{align} \label{eq:naive-deblur}
\underset{z_i \in \R^l}{\text{argmin}} \  \| y - G_\setI(z_i) \|, \quad \quad \hat{i} = G_\setI(\hat{z}_i), 
\end{align}
where we emphasize again that in the optimization program above, the weights of the generator $G_\setI$ are fixed (pretrained). Although non-convex, a local minima $\hat{z}_i$ can be achieved via gradient descent implemented using the back propagation algorithm. The recovered image $\hat{i}$ is obtained by a forward pass of $\hat{z}_i$ through the generative model $G_{\setI}$. Expectedly, this approach fails to produce reasonable recovery results; see Figure \ref{fig:Back-Propagation}. The reason being that this back projection approach completely ignores the knowledge of the forward blur model in \eqref{eq:bid}. 

We now address this shortcoming by including the forward model and blur kernel in the objective (\ref{eq:naive-deblur}). 

\begin{figure*}[t]
\centering
\includegraphics[width=0.82\textwidth]{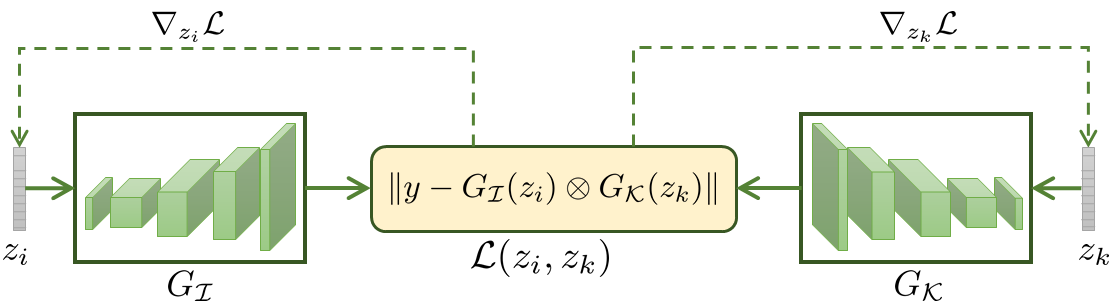}
\caption{\small{Block diagram of proposed approach. Low dimensional parameters $z_i$ and $z_k$ are updated to minimize the measurement loss using alternating gradient descent. The optimal pair $(\hat{z}_i,\hat{z}_k)$ generate image and blur estimates $(G_\setI(\hat{z}_i), G_\setK(\hat{z}_k))$.}}
\label{fig:proposed_approach}
\end{figure*}
\subsection{Deconvolution using Deep Generative Priors} \label{proposedapproach}
We discovered in the previous section that simply finding a clean image close to the blurred one in the range of the image generator $G_{\setI}$ is not good enough. A more natural and effective strategy is to instead find a pair consisting of a clean image and a blur kernel in the range of $G_{\setI}$, and $G_{\setK}$, respectively, whose convolution comes as close to the blurred image $y$ as possible. As outlined in Section \ref{sec:Problem-Formulation}, this amounts to minimizing the measurement loss 
\begin{equation} \label{eq:measurement-loss}
\| y - G_\setI(z_i) \otimes G_\setK(z_k) \|^2,
\end{equation}
over both $z_i$, and $z_k$, where $\otimes$ is the convolution operator.  Incorporating the fact that latent representation vectors $z_i$, and $z_k$ are assumed to be coming from standard Gaussian distributions in both the adversarial learning and variational inference framework, outlined in Section \ref{generative}, we further augment the measurement loss in \eqref{eq:measurement-loss} with $\ell_2$ penalty terms on the latent representations. The resultant optimization program is then 
\begin{equation} \label{eq:regularized-program}
\underset{z_i \in \R^l, z_k \in \R^m}{\text{argmin}}\ \| y - G_\setI(z_i) \otimes G_\setK(z_k) \|^2+ \gamma\| z_i \|^2 + \lambda\| z_k \|^2,
\end{equation}
where $\lambda$, and $\gamma$ are free scalar parameters. For brevity, we denote the objective function above by $\setL(z_i,z_k)$. To  minimize this non-convex objective, we begin by initializing $z_i$, and $z_k$ as standard Gaussian vectors, and then take a gradient step in one of these while fixing the other. To avoid being stuck in a not good enough local minima, we may restart the algorithm with a new random initialization (Random Restarts) when the measurement loss in \eqref{eq:measurement-loss} does not reduce sufficiently after reasonably many iterations. Algorithm \ref{alg:generative-prior-deblurring} formally introduces the proposed alternating gradient descent scheme. Henceforth, we will denote the image deblurred using Algorithm \ref{alg:generative-prior-deblurring} by $\hat{i}_1$.

For computational efficiency, we implement the gradients in the Fourier domain. Define an $n \times n$ DFT matrix $F$ as 
\begin{align*} 
F[\omega,t] = \tfrac{1}{\sqrt{n}}  \mathrm{e}^{-j 2 \pi \omega t/n}, ~ 1 \leq \omega, t \leq n, 
\end{align*}
where $F[\omega,t]$ denotes the $(\omega,t)$th entry of the Fourier matrix. Since the DFT is an isometry, and also diagonalizes the (circular) convolution operator, we can write the loss function in the Fourier domain as 
\begin{equation} \label{eq:loss-fourier}
\setL(z_i,z_k) = 
\| Fy - \sqrt{n}FG_\setI(z_i) \odot 
FG_\setK(z_k) \|^2 + \gamma\| z_i \|^2 + 
\lambda\| z_k \|^2.
\end{equation}

\begin{algorithm}[t]\label{alg:AltGradDescent}
	\caption{Deblurring Strictly under Generative Priors}
	\begin{algorithmic} 
		\STATE \textbf{Input:} $y$, $G_\setI$ and $G_\setK$\\
		\textbf{Output:}  Estimates $\hat{i}_1$ and $\hat{k}$ \\
		\STATE \textbf{Initialize:}\\
		$z_i^{(0)} \sim  \setN(0,I), ~~ z_k^{(0)} \sim  \setN(0,I)$ 
		\FOR{${t = 0,1,2,\ldots, T }$}
		\STATE{$ z_i^{(t+1)} \leftarrow z_i^{(t)}$ - $\eta \nabla_{z_i} \mathcal{L}(z_i^{(t)}, z_k^{(t)})$; } \hspace*{\fill} (\ref{eq:image_grad})\\
		\STATE{$ z_k^{(t+1)} \leftarrow z_k^{(t)}$ - $\eta \nabla_{z_k} \mathcal{L}(z_i^{(t)}, z_k^{(t)})$; } \hspace*{\fill} (\ref{eq:blur_grad})
		\ENDFOR \\
		$\hat{i}_1  \leftarrow G_\setI(z^{(T)}_i), \hat{k}  \leftarrow G_\setK(z^{(T)}_k)$
   \end{algorithmic}
    \label{alg:generative-prior-deblurring}
\end{algorithm}

We now compute the gradient expressions\footnote{ For a function $f(x)$ of variable $x = u+\iota v$, the Wirtinger derivatives of $f(x)$ with respect to $x$, and $\bar{x}$ are defined as 
\begin{align*}
\frac{\partial f}{\partial x} = \frac{1}{2}\bigg(\frac{\partial f}{\partial u}  - \iota \frac{\partial f}{\partial v}  \bigg), \quad \mbox{and} \quad \frac{\partial f}{\partial \bar{x}} = \frac{1}{2}\bigg(\frac{\partial f}{\partial u}  + \iota \frac{\partial f}{\partial v}  \bigg)
\end{align*}} 
for each of the variables $z_i$, and $z_k$. Start by defining residual $r^t$ at the $t$-th iteration: 
\begin{align*}
{r}^{t} := F G_\setK(z_k^{t})\odot \sqrt{n}F G_\setI(z_i^{t})-F y,
\end{align*}
Let $\dot{G}_\setI = \tfrac{\partial}{\partial z_i }{G}_\setI(z_i)$ and $\dot{G}_\setK = \tfrac{\partial}{\partial z_k}{G}_\setK(z_k)$. Then, it is easy to see that 
\begin{align}
&\nabla_{z_i^t} \setL (z_i^t , z_k^t) =  \sqrt{n} \dot{G}_\setI^* F^*\big[{r}^{t}\odot \overline{F {G}_\setK(z_k^{t}})\big] + \gamma z_i^{t},  \label{eq:image_grad} \\
&\nabla_{z_k^t} \setL (z_i^t , z_k^t)  =  \dot{G}_\setK^*F^*\big[{r}^{t}\odot \sqrt{n} \ \overline{F ({G}_\setI(z_i^{t})}\big] + \lambda z_k^{t}.  \label{eq:blur_grad}
\end{align}
For illustration, take the example of a two layer generator ${G}_\setI(z_i)$, which is simply
\[
{G}_\setI(z_i) = \relu(W_2(\relu(W_1 z_i))),
\] 
where $W_1: \R^{l_1 \times l}$, and $W_2: \R^{n \times l_1}$ are the weight matrices at the first, and second layer, respectively. In this case $ \dot{G}_\setI = \widetilde{W}_{2,z_i}\widetilde{W}_{1,z_i}$ where $\widetilde{W}_{1,z_i}=\mbox{diag}(W_1 z_i > 0)W_1$ and $\widetilde{W}_{2,z_i}=\mbox{diag}(W_2\widetilde{W}_{1,z_i} > 0)W_2$.  From this expression, it is clear that alternating gradient descent algorithm for the blind image deblurring requires alternate back propagation through the generators $G_\setI$, and $G_\setK$ as illustrated in Figure \ref{fig:proposed_approach}. To update $z_i^{t-1}$, we fix $z_k^{t-1}$, compute the Fourier transform of a scaling of the residual vector $r^t$, and back propagate it through the generator $G_\setI$. Similar update strategy is employed for $z_k^{t-1}$ keeping $z_i^{t}$ fixed. 

\subsection{Beyond the Range of Generator} \label{sec:proposed-method2}
\begin{algorithm}[t]\label{alg:admm}
	\caption{Deblurring under Classical and Generative Priors}
	\begin{algorithmic} 
		\STATE \textbf{Input:} $y$, $G_\setI$ and $G_\setK$\\
		\textbf{Output:}  Estimates $\hat{i}_2$ and $\hat{k}$ \\
		\STATE \textbf{Initialize:}\\
		$z_i^{(0)} \sim  \setN(0,I), z_k^{(0)} \sim  \setN(0,I), i^{(0)} \sim  \setN(0.5,10^{-2}I)$
		\FOR{${t = 0,1,2,\ldots T}$}
		\STATE{$ z_i^{(t+1)} \leftarrow z_i^{(t)}$ - $\eta \nabla_{z_i} \mathcal{L}(z_i^{(t)}, z_k^{(t)},i^{(t)})$; } \hspace*{\fill} (\ref{eq:opt-admm})\\
		\STATE{$ z_k^{(t+1)} \leftarrow z_k^{(t)}$ - $\eta \nabla_{z_k} \mathcal{L}(z_i^{(t)}, z_k^{(t)},i^{(t)})$; } \hspace*{\fill} (\ref{eq:opt-admm})\\
		\STATE{$ i^{(t+1)} \leftarrow i^{(t)}$ - $\eta \nabla_{i} \mathcal{L}(z_i^{(t)}, z_k^{(t)},i^{(t)})$; } \hspace*{\fill} (\ref{eq:opt-admm})
		\ENDFOR \\
		$\hat{i}_2  \leftarrow {i^{(T)}}, \hat{k}  \leftarrow G_\setK(z_k^{(T)})$
   \end{algorithmic}
    \label{alg:generative+classical-prior-deblurring}
   \label{algorithm:admm}
\end{algorithm}
As described earlier, the optimization program \eqref{eq:regularized-program} implicitly constrains the deblurred image to lie in the range of the generator $G_{\setI}$.  This may leads to some artifacts in the deblurred images when the generator range does not completely span the set $\setI$. This inability of the generator to completely learn the image distribution is often evident in case of more rich and complex natural images. In such cases, it makes more sense to not strictly constrain the recovered image to come from the range of the generator, and rather also explore images a bit outside the range. To accomplish this, we propose minimizing the measurement loss of images inside the range exactly as in \eqref{eq:measurement-loss} together with the measurement loss $\| y - i \otimes G_\setK(z_k) \|^2$ of images not necessarily within the range. The in-range image $G_{\setI}(z_i)$, and the out-range image $i$ are then tied together by minimizing an additional penalty term, $\text{Range Error(i)} := \| i - G_\setI(z_i) \|^2$. The idea is to strictly minimize the range error when pretrained generator has effectively learned the image distribution, and afford some slack when it is not the case. The amount of slack can be controlled by tweaking the  weights attached with each loss term in the final objective. Finally, to guide the search of a best deblurred image beyond the range of the generator, one of the conventional image priors such as total variation measure $\|\cdot\|_{\text{tv}}$ is also introduced. This leads to the following optimization program 

\begin{align} \label{eq:opt-admm}
\underset{i, z_i, z_k}{\text{argmin}}  \ &
\| y - i \otimes G_\setK(z_k) \|^2 + 
\tau \| i - G_\setI(z_i) \|^2 \notag \\
& \quad + \zeta\| y - G_\setI(z_i) \otimes G_\setK(z_k) \|^2 + \rho\|i\|_{\text{tv}}.
\end{align}

All of the variables are randomly initialized, and the objective is minimized using gradient step in each of the unknowns, while fixing the others. The computations of gradients is very similar to the steps outlined in Section \ref{proposedapproach}. We take the solution $\hat{i}$, and $G(z_k)$ as the deblurred image, and the recovered blur kernel. The iterative scheme is formally given in Algorithm \ref{alg:generative+classical-prior-deblurring}. For future references, we will denote the recovered image using Algorithm \ref{alg:generative+classical-prior-deblurring} by $\hat{i}_2$.

\begin{algorithm}[t]
	\caption{Deblurring using Untrained Generative Priors}
	\begin{algorithmic} 
		\STATE \textbf{Input:} $y$, $G_\setI$ and $G_\setK$\\		
		\textbf{Output:}  Estimates $\hat{i}_3$ and $\hat{k}$ \\
		\STATE \textbf{Initialize:}\\
		$z_i^{(0)} \sim  \setN(0,I), z_k^{(0)} \sim  \setN(0,I), W \sim  \setN(0,I)$ 
		\STATE {$W^{(0)} \leftarrow \underset{W}{\text{argmin}} \quad \Vert y - G_\setI(z_i^{(0)},W)\Vert^2$}
		\\
		\FOR{${t = 0,1,2,\ldots }$}
		\STATE{$ z_i^{(t+1)} \leftarrow z_i^{(t)}$ - $\eta \nabla_{z_i} \setL(z_i^{(t)}, z_k^{(t)}, W^{(t)})$; } \hspace*{\fill} (\ref{eq:dpdn-loss})\\
		\STATE{$ z_k^{(t+1)} \leftarrow z_k^{(t)}$ - $\eta \nabla_{z_k} \setL(z_i^{(t)}, z_k^{(t)}, W^{(t)})$; } \hspace*{\fill} (\ref{eq:dpdn-loss})\\
		\STATE{$ W^{(t+1)} \leftarrow W^{(t)}$ - $\eta \nabla_{W} \setL(z_i^{(t)}, z_k^{(t)}, W^{(t)})$; }   \hspace*{\fill} (\ref{eq:dpdn-loss})
		\ENDFOR \\
		$\hat{i}_3  \leftarrow G_\setI(z_i^{(T)},W^{(T)}), \hat{k}  \leftarrow G_\setK(z_k^{(T)})$
   \end{algorithmic}
   \label{alg:dpd-n}
\end{algorithm}

\subsection{Untrained Generative Priors} \label{proposed-method3}

As will be shown in the numerics below that the pretrained generative models effectively regularize the deblurring and produce competitive results, however, convincing performance is limited to the image datasets such as faces, and numbers, etc. that are somewhat effectively learned by the generative models. In comparison, on more complex/rich, and hence not effectively learned image datasets such as natural scenery images, the regularization ability of generative models is expected to suffer. This discussion begs a question: can only a pretrained generator act as a good image prior in the deblurring inverse problem? The answer to this question is surprisingly, no; our experiments suggest that even an untrained structured generative network acts as a good prior for natural images in deblurring. Similar observation was first made in \cite{ulyanov2017deep} in other image restoration contexts. This surprising observation suggests that the structure (deep convolutional layers) of the generative network alone (without any pretraining) captures some of the image statistics, and hence can act as a reasonable prior in the inverse problem. Of course, this untrained generative network is not as effective a prior as a pretrained one. However, importantly for us, this ability of a deep convolutional network makes the case for continuing to employ it as a prior on complex images on which the generator is either  \textit{not well trained} or even untrained. 

We will continue to use the easy to train generator (slim network) for blurs as a pretrained network while the weights of the untrained image generator will be updated together with the input vectors $z_i$, and $z_k$ to minimize the measurement loss. The deblurring scheme previously was concerned with only updating $z_i$, and $z_k$. Importantly, unlike the pretrained image generator; trained on thousands of image examples, the weights of the untrained generator are learned on one blurred image $y$ only in the deblurring process itself. To encourage a sane weight update (leading to realistic generated images), we add a total variation ($\|\cdot\|_{\text{tv}}$) penalty on the output of the image generator. This assists the generator to learn weights and produce natural images that typically have a smaller tv measure (piecewise constant). Just as before  \eqref{eq:regularized-program}, we also add $\ell_2$ penalty on $z_k$. The resultant optimization program for image deblurring in this case is 
\begin{align}\label{eq:dpdn-loss}
\underset{z_i,z_k,W}{\text{argmin}} \ &\| y - G_\setI(z_i,W) \otimes G_\setK(z_k) \|^2 + \kappa \| z_k \|^2\notag\\
&\qquad  +\nu \|G_\setI(z_i,W)\|_{\text{tv}},
\end{align}
where $G_\setI(z_i,W)$ denotes an image generator with weight parameters $W$, and  input $z_i$. We minimize the objective $\setL(z_i,z_k,W)$ in the optimization program above by alternatively taking gradient steps in each of the unknowns while fixing the others. The vectors $z_i$, and $z_k$ are initialized as random Gaussain vectors. We initialize the weights $W$ of $G_{\setI}$ by fitting $G_{\setI}(z_i,W)$ to the given blurry image $y$ for a fixed random input $z_i$. This is equivalent to solving the optimization program below
\begin{equation}\label{eq:deep-prior-weights}
W^* = \underset{W}{\text{argmin}} \quad \Vert y - G_\setI(z_i,W)\Vert^2. 
\end{equation}
Formally, the iterative scheme to minimize the optimization program in \eqref{eq:dpdn-loss} is given in Algorithm \ref{alg:dpd-n}. From the minimizer $(\hat{z}_i,\hat{z}_k,\hat{W})$, the desired deblurred image, and the blur kernel are obtained using a forward pass as $\hat{i}_3 = G_\setI(\hat{z}_i,\hat{W})$, and $\hat{k} =  G_\setK(\hat{z}_k)$, respectively.


\section{Experimental Results} \label{experiments}
We now provide a comprehensive set of experiments to evaluate the performance of proposed novel deblurring approach under generative priors. For brevity, notations have been introduced in Table \ref{table:notation}. We begin by giving a description of the clean image and blur datasets, and a brief mention of the corresponding pretrained generative models for each dataset in Section \ref{sec:Datasets+GenerativeModels}. A description of the baseline methods for deblurring is provided in Section \ref{sec:BaselineMethods}. Section \ref{sec:Exps-PretrainedPriors} gives a detailed qualitative, and quantitative performance evaluations of our proposed techniques in comparison to the baseline methods. The choice of free parameters for both Algorithm \ref{alg:generative-prior-deblurring} and \ref{alg:generative+classical-prior-deblurring} are mentioned in Table \ref{table:alg1-param} and \ref{table:alg2-param}, respectively. We also evaluate performance under increasing noise and large blurs. In addition, we discuss the impact of increasing the latent dimension, and multiple random restarts in the proposed algorithm on the deblurred images. Section \ref{sec:Exps-UntrainedPriors} showcases the image deblurring results on complex natural images using untrained generative priors. In all experiments, we use noisy blurred images, generated by convolving images $i$, and blurs $k$ from their respective test sets and adding 1$\%$ \footnote{For an image scaled between 0 and 1, Gaussian noise of $1\%$ translates to Gaussian noise with standard deviation $\sigma = 0.01$ and mean $\mu=0$.} Gaussian noise (unless stated otherwise).

\begin{figure}[t]
\centering
\includegraphics[width=\columnwidth]{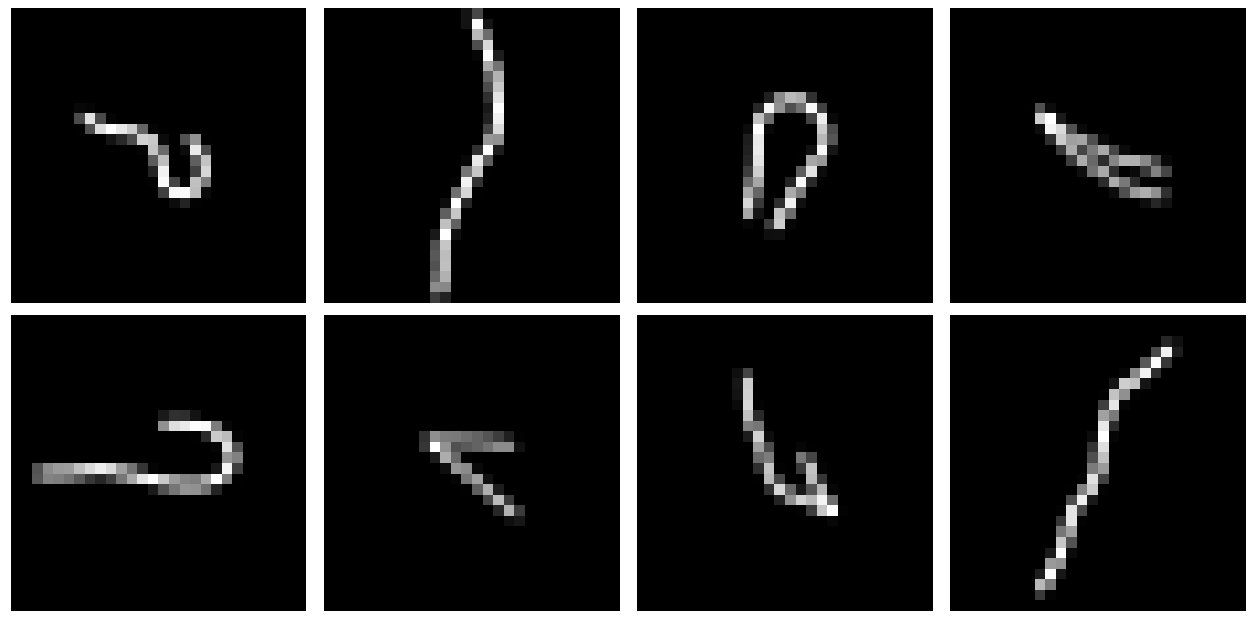}
\caption{\small  Synthetically generated blur kernels.}
\label{fig:training_blurs}
\end{figure}
\newcolumntype{P}[1]{>{\centering\arraybackslash}p{#1}}
\begin{table}[]
\centering
\renewcommand{\arraystretch}{1.4}
\scalebox{0.86}{
\begin{tabular}{|m{2.2cm}|P{3.5cm}|m{0.75cm}|m{0.75cm}|m{0.75cm}|}
\hline
\multicolumn{1}{|c|}{\multirow{2}{*}{\textbf{Input}}} & \multicolumn{1}{c|}{\multirow{2}{*}{\textbf{Description}}} & \multicolumn{3}{c|}{\textbf{Output of Algorithm}} \\ \cline{3-5} 
\multicolumn{1}{|c|}{} & \multicolumn{1}{c|}{} & \textbf{Alg-1} & \textbf{Alg-2} & \textbf{Alg-3} \\ \hline
$y = i_{\text{test}} \otimes k + n $ & Blurry image generated from test set images & $\hat{i}_{1}$ & $\hat{i}_{2}$ & $\hat{i}_{3}$ \\ \hline
$y = i_{\text{sample}} \otimes k + n $ & Blurry image generated from sampled image $i_{\text{sample}}=G_{\mathcal{I}}(z_i)$ $z_i = \setN(0,I)$ & $\hat{i}_{\text{sample}}$ & -- & -- \\ \hline
$y = i_{\text{range}} \otimes k + n$ & Blurry image generated from closest image to $i_{\text{test}}$ in range of $G_\setI$ & $\hat{i}_{\text{range}}$ & -- & -- \\ \hline
\end{tabular}
}
\caption{Notations developed for different images used to generate blurry images $y$ for deblurring using proposed algorithms and their corresponding outputs.}
\renewcommand{\arraystretch}{1.0}
\label{table:notation}
\end{table}

\subsection{Datasets and Generative Models} \label{sec:Datasets+GenerativeModels}
To evaluate the proposed technique, we choose three image datasets. First dataset, SVHN, consists of house number images from Google street view. A total of 531K images, each of dimension $32\times32 \times 3$, are available in SVHN out of which 30K are held out as test set. Second dataset, Shoes \cite{yu2014fine} consists of 50K RGB examples of shoes, resized to $64 \times 64 \times 3$. We leave $1000$ images for testing and use the rest as training set. Third dataset, CelebA, consists of relatively more complex images of celebrity faces. A total of 200K, each center cropped to dimension $64 \times 64 \times 3$, are available out of which 22K are held out as a test set.

A motion blur dataset is generated consisting of small to very large blurs of lengths varying between 5 and 28; following strategy given in \cite{boracchi2012modeling}. Some of the representative blurs of this dataset are shown in Figure \ref{fig:training_blurs}. We generate 80K blurs out of which 20K is held out as a test set.

The generative model of SVHN images is a trained VAE with the network architecture described in Table \ref{table:vae-architectures}. The dimension of the latent space of VAE is 100, and training is carried out on SVHN with a batch size of 1500, and a learning rate of $10^{-5}$ using the Adam optimizer. After training, the decoder part is extracted as the desired generative model $G_{\setI}$. For Shoes and CelebA, the generative model $G_{\setI}$ is the default deep convolutional generative adversarial network (DCGAN)of \cite{salimans2016improved}.

The generative model of motion blur dataset is a trained VAE  with the network architecture given in Table \ref{table:vae-architectures}. This VAE is trained using Adam optimizer with latent dimension 50, batch size 5, and learning rate $10^{-5}$. After training, the decoder part is extracted as the desired generative model $G_{\setK}$.

\subsection{Baseline Methods}\label{sec:BaselineMethods}
Among the conventional algorithms using engineered priors, we choose dark prior (DP) \cite{pan2016blind}, extreme channel prior (EP) \cite{yan2017image}, outlier handling (OH) \cite{dong2017blind}, and learned data fitting (DF) \cite{pan2017learning} based blind deblurring as baseline algorithms. We optimized the parameters of these methods in each experiment to obtain the best possible baseline results. Out of the more recent, and very competitive data driven approaches for deblurring, we choose \cite{hradivs2015convolutional} that trains a convolutional neural network (CNN) in an end-to-end manner,  and \cite{kupyn2017deblurgan} that trains a neural network (DeblurGAN) in an adversarial manner. Each of these networks is trained on SVHN, and CelebA. For CNN, we train a slightly modified (fine-tuned) version of \cite{hradivs2015convolutional} using Adam optimizer with learning rate $10^{-4}$ and batch size 16. To train the DeblurGAN, we use the code provided by authors of \cite{kupyn2017deblurgan}. Deblurred images from these baseline methods will be referred to as $i_\text{DP}$, $i_\text{EP}$, ${i_\text{OH}}$, ${i_\text{DF}}$, $i_\text{CNN}$ and $i_\text{DeGAN}$.

\begin{table}[t]
\centering
\renewcommand{\arraystretch}{1.3}
\begin{tabular}{
|>{\centering\arraybackslash}p{0.9cm}
|>{\centering\arraybackslash}p{0.44cm}
|>{\centering\arraybackslash}p{0.44cm}
|>{\centering\arraybackslash}p{1cm}
|>{\centering\arraybackslash}p{2.0cm}
|>{\centering\arraybackslash}p{1.0cm}|}
\hline
Dataset & $\lambda$ & $\gamma$ & Steps(t) & Step Size & Random Restarts \\ \hline
SVHN & 0.01 & 0.01 & 6,000 & $0.01\exp^{-\frac{t}{1000}}$ & 10 \\ \hline
Shoes & 0.01 & 0.01 & 10,000 & $0.01\exp^{-\frac{t}{1000}}$ & 10 \\ \hline
CelebA & 0.01 & 0.01 & 10,000 & $0.01\exp^{-\frac{t}{1000}}$ & 10 \\ \hline
\end{tabular}
\renewcommand{\arraystretch}{1.0}
\caption{Algorithm 1 Parameters.}
\label{table:alg1-param}
\end{table}

\begin{table}[t]
\centering
\renewcommand{\arraystretch}{1.3}
\begin{tabular}{
|>{\centering\arraybackslash}p{0.9cm}
|>{\centering\arraybackslash}p{0.44cm}
|>{\centering\arraybackslash}p{0.44cm}
|>{\centering\arraybackslash}p{0.51cm}
|>{\centering\arraybackslash}p{1cm}
|>{\centering\arraybackslash}p{1.3cm}
|>{\centering\arraybackslash}p{1.0cm}|}
\hline
Dataset & $\tau$ & $\zeta$ & $\rho$ & Steps(t) & Step Size & Random Restarts \\ \hline
Shoes & 100 & 0.5 & $10^{-3}$ & 10,000 & 0.005 (adam) & 10 \\ \hline
CelebA & 100 & 0.5 & $10^{-3}$ & 10,000 & 0.005 (adam) & 10 \\ \hline
\end{tabular}
\renewcommand{\arraystretch}{1.0}
\caption{Algorithm 2 Parameters.}
\label{table:alg2-param}
\end{table}

\begin{table*}
\centering
\scalebox{0.9}{
\setlength{\arrayrulewidth}{.1mm}
\renewcommand{\arraystretch}{1.3}
\begin{tabular}{ |c|>{\centering\arraybackslash}p{7cm}|>{\centering\arraybackslash}p{7cm}| }
\hline
\multicolumn{3}{|c|}{\textbf{Model Architectures}} \\
\hline
\textbf{Model} & \textbf{Encoder} & \textbf{Decoder} \\ 
\hline
\begin{tabular}[c]{@{}l@{}}\\Blur VAE\end{tabular} & conv($20$, $2\times2$, $1$) $\rightarrow$ relu $\rightarrow$ maxpool($2\times2$, $2$) $\rightarrow$ conv($20$, $2\times2$, $1$) $\rightarrow$ relu $\rightarrow$ maxpool($2\times2$, $2$) $\rightarrow$ fc($50$), fc($50$) $\rightarrow$ $z_k$ 
 & $z_k$ $\rightarrow$ fc($720$) $\rightarrow$ relu $\rightarrow$ reshape $\rightarrow$ upsample($2\times2$) $\rightarrow$ convT($20$, $2\times2$, $1$) $\rightarrow$ relu $\rightarrow$ upsample($2\times2$) $\rightarrow$ convT($20$, $2\times2$, $1$) $\rightarrow$ relu $\rightarrow$ convT($1$, $2\times2$, $1$) $\rightarrow$ relu   \\[0.5ex] 
\hline 
\begin{tabular}[c]{@{}l@{}} \\SVHN VAE\end{tabular} & conv($128$, $2\times2$, $2$) $\rightarrow$ batch-norm $\rightarrow$ relu $\rightarrow$ conv($256$, $2\times2$, $2$) $\rightarrow$ batch-norm $\rightarrow$ relu $\rightarrow$ conv($512$, $2\times2$, $2$) $\rightarrow$ batch-norm $\rightarrow$ relu $\rightarrow$ fc($100$), fc($100$) $\rightarrow$ $z_i$
 & $z_i$ $\rightarrow$ fc($8192$) $\rightarrow$ reshape $\rightarrow$ convT($512$, $2\times2$, $2$) $\rightarrow$ batch-norm $\rightarrow$ relu $\rightarrow$ convT($256$, $2\times2$, $2$) $\rightarrow$ batch-norm $\rightarrow$ relu $\rightarrow$ convT($128$, $2\times2$, $2$) $\rightarrow$ batch-norm $\rightarrow$ relu $\rightarrow$ conv($3$, $1\times1$, $1$) $\rightarrow$ sigmoid  \\
\hline
\end{tabular}}
\renewcommand{\arraystretch}{1.0}
\captionsetup{width=.9\textwidth}
\caption{ \small{ Architectures for VAEs used for Blur and SVHN. Here, conv($m$,$n$,$s$) represents convolutional layer with $m$ filters of size $n$ and stride $s$. Similarly, convT represents transposed convolution layer. Maxpool($n$,$m$) represents a max pooling layer with stride $m$ and pool size of $n$. Finally, fc($m$) represents a fully connected layer of size $m$. The decoder is designed to be a mirror reflection of the encoder in each case.}}
\label{table:vae-architectures}
\end{table*}

\subsection{Deblurring Results under Pretrained Generative Priors}\label{sec:Exps-PretrainedPriors}

We now evaluate the performance of Algorithm \ref{alg:generative-prior-deblurring} under small to heavy blurs, and varying degrees of additive measurement noise. As will be shown, both qualitatively and quantitatively, that Algorithm \ref{alg:generative-prior-deblurring} produces encouraging deblurring results, especially, under large blurs, and heavy noise. However, the central limiting factor in the performance is the ability of the generator to represent the (original, clean) image to be recovered. As pointed out earlier that often the generators are not fully expressive (cannot generate new representative samples) on a rich/complex image class such as face images compared to a compact/simple image class such as numbers. Such a generator mostly cannot \textit{adequately} represent a new image in its range. Since Algorithm \ref{alg:generative-prior-deblurring} strictly constrains the recovered image to lie in the range of image generator, its performance depends on how well the range of the generator spans the image class. Given an arbitrary test image $i_{\text{test}}$ in the set $\setI$, the closest image $i_{\text{range}}$, in the range of the generator, to $i_{\text{test}}$ is computed by solving the following optimization program
\begin{align*}
z_{\text{test}} := \underset{z} {\text{argmin}}  \|i_{\text{test}}-G_{\setI}(z)\|^2,  \quad i_{\text{range}} = G_{\setI}(z_{\text{test}})
\end{align*}
We solve the optimization program by running $10,000(6,000)$ gradient descent steps with a step size of $0.001(0.01)$ for CelebA(SVHN). Parameters for Shoes are the same as CelebA.

A more expressive generator leads to a better deblurring performance as it can well represent an arbitrary original (clean) image $i_{\text{test}}$ leading to a smaller mismatch  
 \begin{align}\label{eq:range_error}
 \text{range error} := \|i_{\text{test}} - i_{\text{range}}\|,
 \end{align}
  to the corresponding range image $i_{\text{range}}$. Using the triangle inequality, we have the following upper bound on the overall recovery error $\|\hat{i}-i_{\text{test}}\|$ between the deblurred image $\hat{i}$, and true image $i_{\text{test}}$ in terms of the range error.
\begin{align}\label{eq:overall-error}
 	\text{overall error}:= \|\hat{i}-i_{\text{test}}\| \leq \|\hat{i}-i_{\text{range}}\| + \| i_{\text{range}}-i_{\text{test}}\|.
\end{align}

\begin{figure}[t]
\centering
\includegraphics[width=\columnwidth]{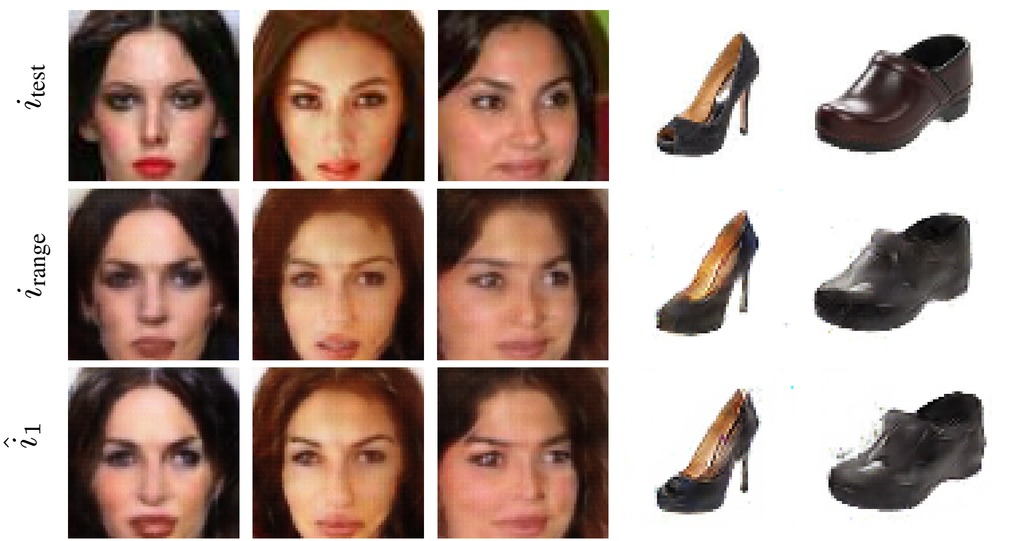}
\caption{\small  Generator Range Analysis. This figure visually demonstrates that for each test image when blurred, Algorithm \ref{alg:generative-prior-deblurring} tends to recover corresponding range image. For shoes, $\hat{i}_1$ fails to capture texture of $i_{\text{test}}$ similar to $i_\text{range}$. Faces images show this more clearly as deblurred images are only semantically different from range images.}
\label{fig:range-images}
\end{figure}

\subsubsection{Impact of Generator Range on Image Deblurring} The range error purely depends on the expressive power of the generator that in turn relies on factors, such as training scheme, network structure and depth, clearly determined by the available computational resources. Therefore, to judge the deblurring algorithms independently of generator limitations, we present their deblurring performance on range image $i_{\text{range}}$;  we do this by generating a blurred image $y = i_{\text{range}} \otimes k + n$ from an image $i_{\text{range}}$ already in the range of the generator; this implicitly removes the range error in \eqref{eq:overall-error} as now $i_{\text{test}} = i_{\text{range}}$. We call this range image deblurring, and specifically the deblurred image is obtained using Algorithm \ref{alg:generative-prior-deblurring}, and is denoted by $\hat{i}_{\text{range}}$. For completeness, we also assess the overall performance of the algorithm by deblurring arbitrary blurred images $y = i_{\text{test}} \otimes k +n$, where $i_{\text{test}}$ is not necessarily in the range of the generator. Unlike above, the overall error in this case accounts for the range error as well. We call this arbitrary image deblurring, and specifically the deblurred image is obtained using Algorithm \ref{alg:generative-prior-deblurring}, and is denoted by $\hat{i}_1$. Figure \ref{fig:range-images} shows a qualitative comparison between $i_{\text{test}}$, $i_{\text{range}}$, and $\hat{i}_1$ on CelebA and Shoes dataset. It is clear that the recovered image $\hat{i}_1$ is a good approximation of the range image $i_{\text{range}}$, \textit{closest} to the original (clean) image $i_{\text{test}}$ in the range of $G_\setI$. Evidently, the deviations of $i_{\text{range}}$ in referenced figure from $i$ indicate the limitation of the used image generative network. There can be many suspects that contribute to this range issue of the generator; mode collapse being the most likely \cite{metz2016unrolled}. Currently alot of work is being done to resolve mode collapse and other issues in GANs \cite{mukherjee2018clustergan, thanh2018catastrophic, srivastava2017veegan}, therefore a better generative model with more expressive range will undoubtedly perform better.

\begin{figure*}[t]
\centering
\includegraphics[width=\textwidth]{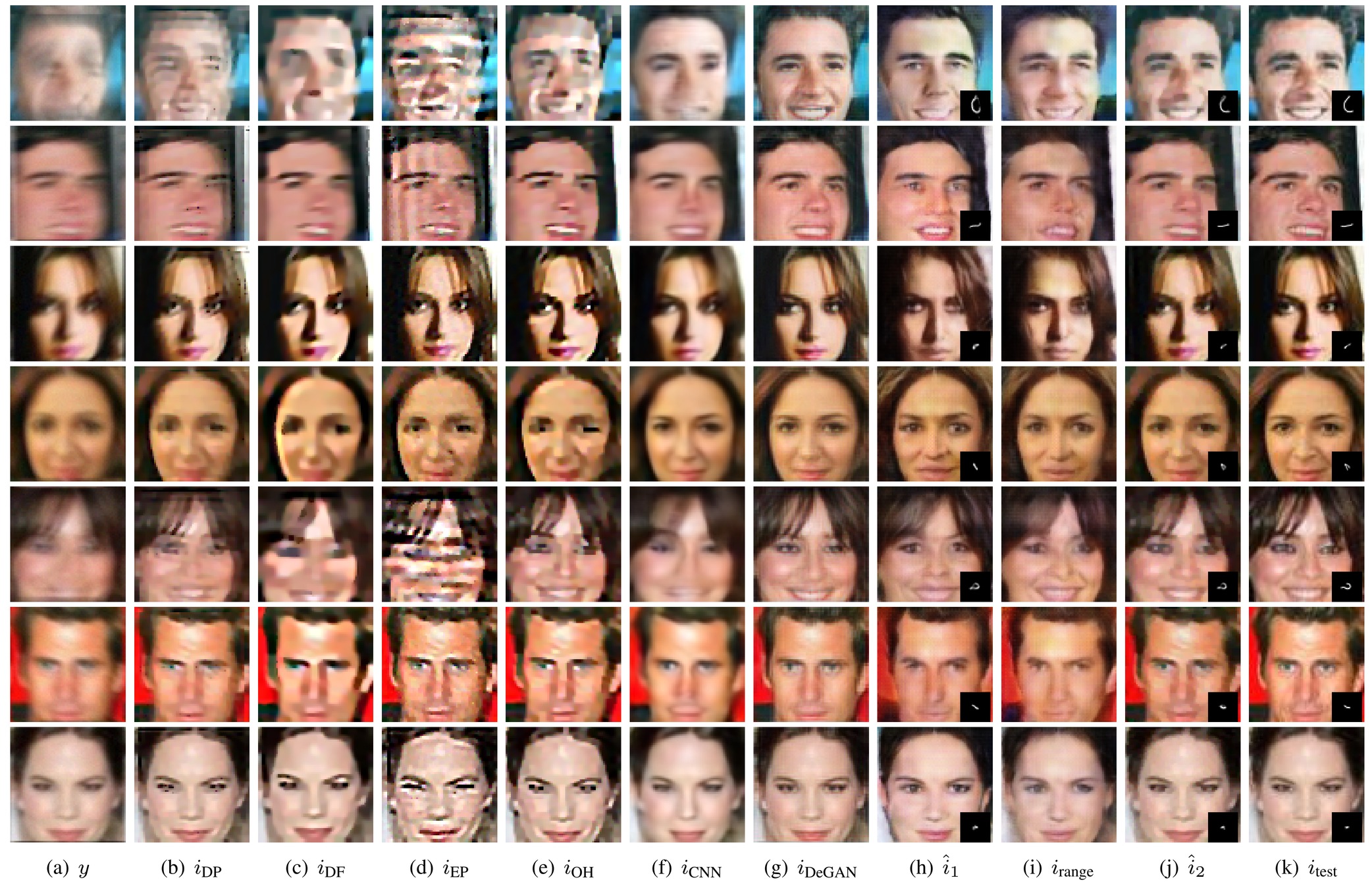}
\caption{\small{Image deblurring results on CelebA using Algorithm \ref{alg:generative-prior-deblurring} and \ref{alg:generative+classical-prior-deblurring}. It can be seen that $\hat{i}_1$ is in close resemblance to $i_{\text{range}}$ (closest image in the generator range to the original image), where as $\hat{i}_2$ is almost exactly $i_{\text{test}}$, thus mitigating the range issue of image generator.}}
\label{fig:celebA-results}
\end{figure*}

Algorithm \ref{alg:generative+classical-prior-deblurring} mitigates the range error by not strictly constraining the recovered image to lie in the range of the image generator, and uses a combination of the generative prior, and a classical engineered prior; for details, see  Section \ref{sec:proposed-method2}. The blurred image in this case is again $y = i_{\text{test}} \otimes k + n$ for an arbitrary (not necessarily in the range) image $i_{\text{test}}$ in $\setI$.  The image deblurred using Algorithm \ref{alg:generative+classical-prior-deblurring} is denoted as $\hat{i}_2$. For comparison, we present the recovered images using this approach in Figure \ref{fig:celebA-results}. It can be seen, again, that $\hat{i}_1$ is in close resemblance to $i_{\text{range}}$, where as $\hat{i}_2$ is almost exactly $i_{\text{test}}$, thus mitigating the range issue of the generator $G_\setI$.

\begin{figure*}
\centering
\includegraphics[width=\textwidth]{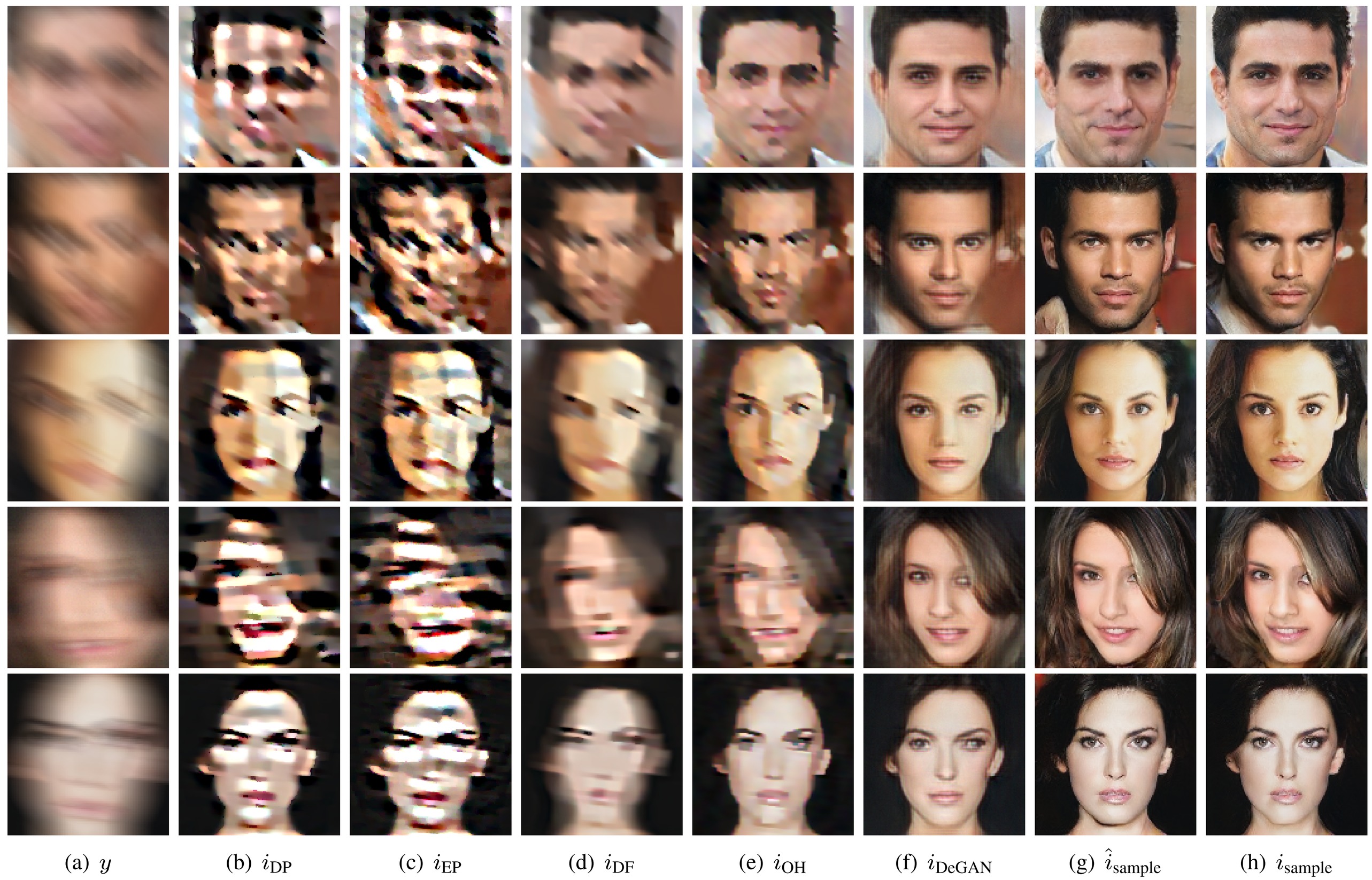}
\caption{\small{Image deblurring results on blurry images generated from samples, $i_\text{sample}$, of PG-GAN using Algorithm \ref{alg:generative-prior-deblurring}. It can be seen that visually appealing images are recovered, $\hat{i}_\text{sample}$, from blurry ones.}}
\label{fig:pg-gan-results}
\end{figure*}

\begin{figure}[h]
\centering
\includegraphics[width=\columnwidth]{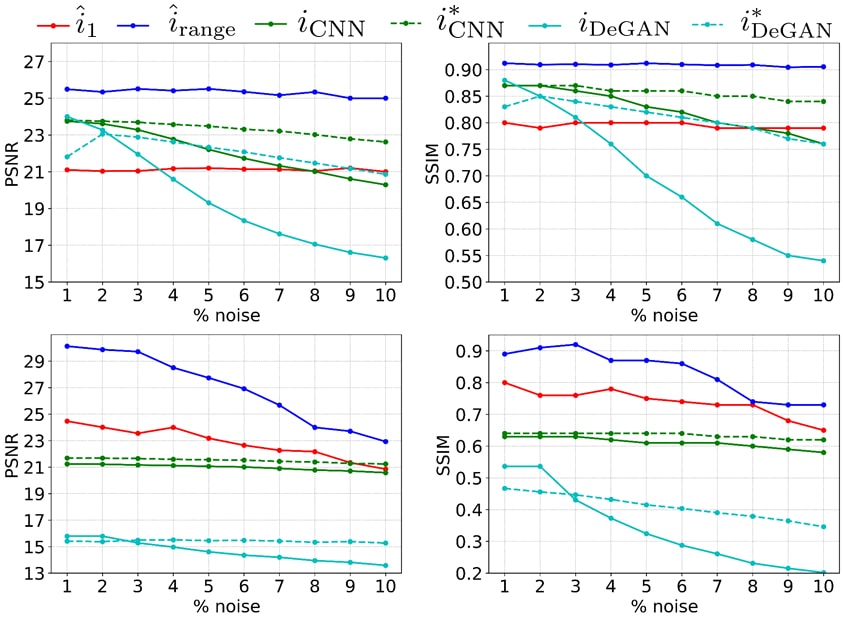}
\caption{\small Noise Analysis. Performance of our methods on CelebA (first row) and SVHN (second row) with increasing noise levels for both range and test images against baseline methods. * indicates that these models were trained on $1-10$\% noise levels.}
\label{fig:psnr-ssim-noise}
\end{figure}

\begin{figure}[h]
\centering
\includegraphics[width=\columnwidth]{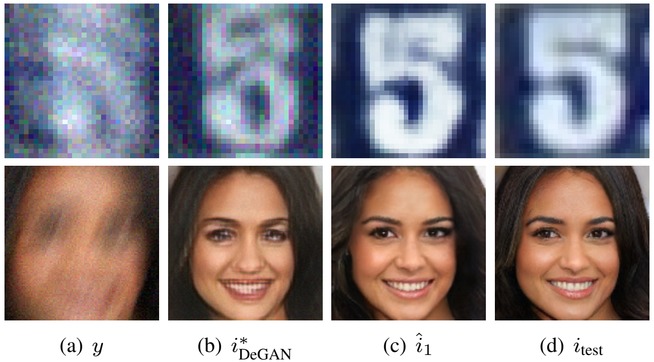}
\caption{\small Visual Comparison of DeblurGAN ($i_\text{DeGAN}^*$) trained on $1$to$10\%$ noise with Algorithm \ref{alg:generative-prior-deblurring} on noisy images from SVHN (top row) and samples from PG-GAN (bottom row). }
\label{fig:results-heavynoise}
\end{figure}

\begin{figure*}[h]
\centering
\includegraphics[width=\textwidth]{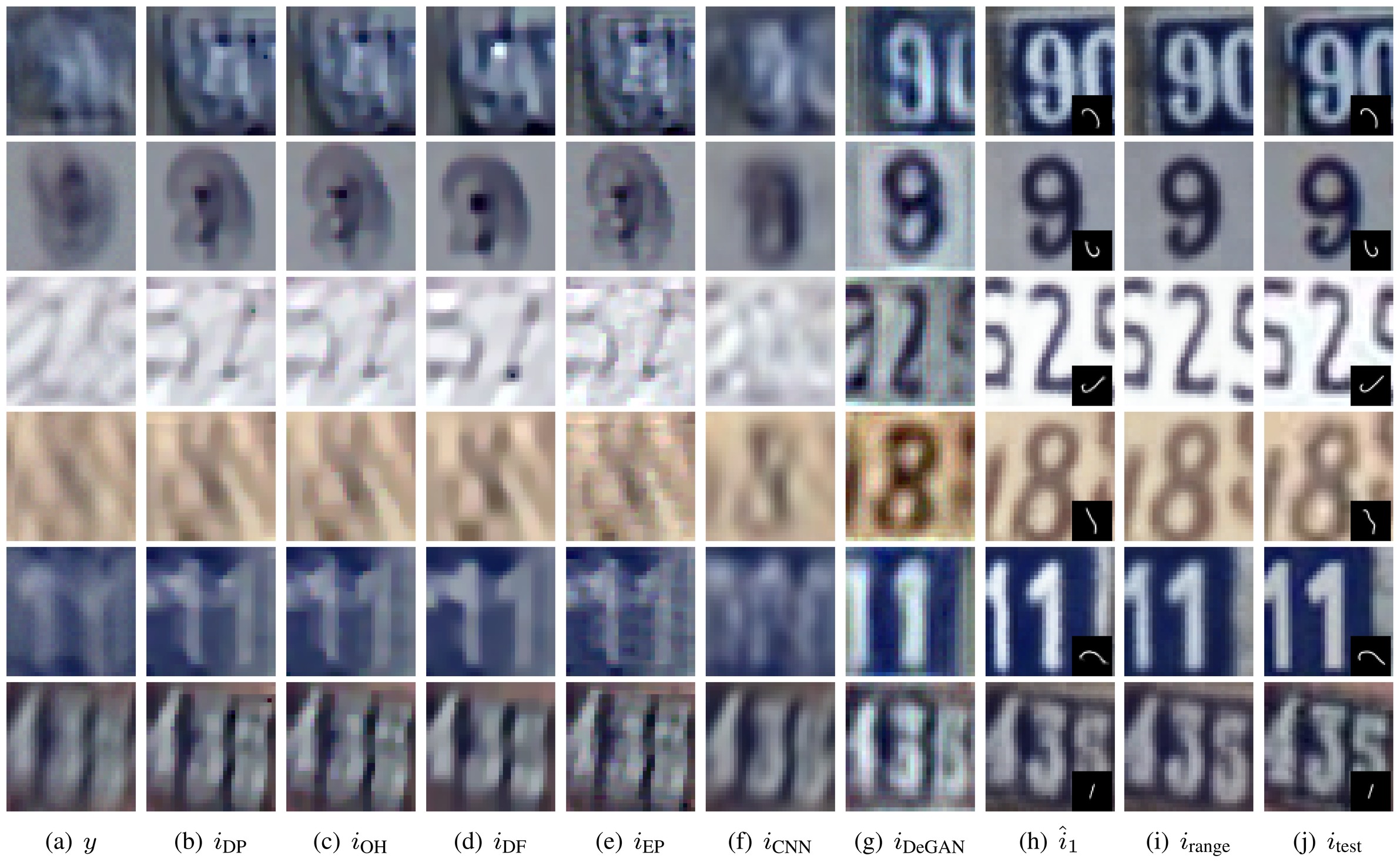}
\caption{\small{Image deblurring results on SVHN images using Algorithm \ref{alg:generative-prior-deblurring}. It can be seen that due to the simplicity of these images, $\hat{i}_1$ is a visually a very good estimate of $i_{\text{test}}$, due to the close proximity between $i_\text{range}$ and $i_{\text{test}}$.}}
\label{fig:svhn-results}
\end{figure*}
 
\subsubsection{Qualitative Results on CelebA}
Figure \ref{fig:celebA-results} gives a qualitative comparison between $i$, $i_{\text{range}}$, $\hat{i}_1$, and $\hat{i}_2$ on CelebA dataset. We also show the image deblurring using the baseline methods introduced in Section \ref{sec:BaselineMethods}.  Unfortuanately, the deblurred images under engineered priors are qualitatively a lot inferior than the deblurred images $\hat{i}_1$, and $\hat{i}_2$ under the proposed generative priors, especially under large blurs. On the other hand, the end-to-end training based approaches CNN, and DeblurGAN perform relatively better, however, the well reputed CNN is still displaying over smoothed images with missing edge details, etc  compared to our results $\hat{i}_2$. DeblurGAN, though competitive, is outperformed by the proposed Algorithm \ref{alg:generative+classical-prior-deblurring} by more than 1.5dB. On closer inspection, $i_{\text{DeGAN}}$ although sharp, deviates from $i_{\text{test}}$, whereas $\hat{i}_2$ tends to agree more closely with $i_{\text{test}}$. A close comparison between  the recovered images $\hat{i}_1$, and $\hat{i}_2$ reveals that later often performs better than former. The images $\hat{i}_1$ are sharp and with well defined facial boundaries and markers owing to the fact they strictly come from the range of the generator, however, in doing so these images might end up changing some image features such as expressions, nose, etc. On a close inspection, it becomes clear that how well $\hat{i}_1$ approximates $i_{\text{test}}$ roughly depends (see, images in the second row specifically of Figure \ref{fig:range-images}) on how close $i_{\text{range}}$ is to $i_{\text{test}}$ exactly, as discussed at length in the beginning of this section.   While as $\hat{i}_2$ are allowed some leverage, and are not strictly confined to the range of the generator, they tend to agree more closely with the ground truth. We go on further by utilizing pretrained PG-GAN \cite{karras2017progressive} in our algorithm by convolving sampled images with large blurs; see Figure \ref{fig:pg-gan-results}. It has been observed that pre-trained generators struggle at higher resolutions \cite{athar2018latent}, so we restrict our results at $128 \times 128$ resolution. We skip the discussion of PG-GAN over test set, as we observed that PG-GAN did not generalize well to test set; mode collapse being the likely suspect. In Figure \ref{fig:pg-gan-results}, it can be seen that under expressive generative priors our approach exceeds all other baseline methods recovering fine details from extremely blurry images.

 \subsubsection{Qualitative Results on SVHN}
Figure \ref{fig:svhn-results} gives qualitative comparison between proposed and baseline methods on SVHN dataset. Here the deblurring under classical priors again clearly under performs compared to the proposed image deblurring results $\hat{i}_1$. CNN also continues to be inferior, and the DeblurGAN that produced competitive results on CelebA and Shoes above also shows artifacts. We do not include the results $\hat{i}_2$ in these comparison as $\hat{i}_1$ already comprehensively outperform the other techniques on this dataset. The convincing results $\hat{i}_1$ are a manifestation of the fact that unlike the relatively complex CelebA and Shoes datasets, the simpler image dataset SVHN is effectively spanned by the range of the image generator. 

\begin{table}[t]
\centering
\scalebox{0.9}{
\renewcommand{\arraystretch}{1.2}
\begin{tabular}{|l|r|r|r|r|r|r|}
\hline
\multicolumn{1}{|c|}{\multirow{2}{*}{\textbf{Method}}} & \multicolumn{2}{c|}{\textbf{SVHN}} & \multicolumn{2}{c|}{\textbf{Shoes}} & \multicolumn{2}{c|}{\textbf{CelebA}} \\ \cline{2-7} 
\multicolumn{1}{|c|}{} & \multicolumn{1}{c|}{\textbf{PSNR}} & \multicolumn{1}{c|}{\textbf{SSIM}} & \multicolumn{1}{c|}{\textbf{PSNR}} & \multicolumn{1}{c|}{\textbf{SSIM}} & \multicolumn{1}{c|}{\textbf{PSNR}} & \multicolumn{1}{c|}{\textbf{SSIM}} \\ \hline
$i_\text{EP}$ \cite{yan2017image} & 20.35 & 0.55 & 18.33 & 0.73 & 17.80 & 0.70 \\ \hline
$i_\text{DF}$ \cite{pan2017learning} & 20.64 & 0.60 & 17.79 & 0.73 & 20.00 & 0.79 \\ \hline
$i_\text{OH}$ \cite{dong2017blind} & 20.82 & 0.58 & 19.04 & 0.76 & 20.71 & 0.81 \\ \hline
$i_\text{DP}$ \cite{pan2016blind} & 20.91 & 0.58 & 18.45 & 0.74 & 21.09 & 0.79 \\ \hline
$i_\text{DeGAN}$ \cite{kupyn2017deblurgan} & 15.79 & 0.54 & 21.84 & 0.85  & 24.01 & 0.88 \\ \hline
$i_\text{CNN}$ \cite{hradivs2015convolutional} & 21.24 & 0.63 & 24.76 & 0.89 & 23.75 & 0.87 \\ \hline
$\hat{i}_1$ & \textbf{24.47} & \textbf{0.80} & 21.20 & 0.83 & 21.11 & 0.80 \\ \hline
$\hat{i}_2$ & - & - & \textbf{26.98} & \textbf{0.93} & \textbf{26.60} & \textbf{0.93} \\ \hline
$\hat{i}_\text{range}$ & 30.13 & 0.89 & 23.93 & 0.87 & 25.49 & 0.91 \\ \hline
\end{tabular}
\renewcommand{\arraystretch}{1.0}
}
\caption{ \small{Quantitative comparison of proposed approach with baseline methods on CelebA, SVHN, and Shoes dataset. Table shows average PSNR and SSIM on 80 random images from respective test sets.}}
\label{table:psnr-ssim-results}
\end{table}

\subsubsection{Quantitative Results}
Quantitative results for CelebA, Shoes\footnote{For qualitative results, see supplementary material.} and SVHN using peak-signal-to-noise ratio (PSNR) and structural-similarity index (SSIM) \cite{wang2004image}, averaged over 80 test set images, are given in Table \ref{table:psnr-ssim-results}. On CelebA and Shoes, the results clearly show a better performance of our proposed Algorithm 2, on average, compared to all baseline methods. On SVHN, the results show that Algorithm \ref{alg:generative-prior-deblurring} outperforms all competitors. The fact that Algorithm \ref{alg:generative-prior-deblurring} performs more convincingly on SVHN is explained by observing that the range images $i_{\text{range}}$ in SVHN are quantitatively much better compared to range images of CelebA and Shoes.

\subsubsection{Robustness against Noise}
Figure \ref{fig:psnr-ssim-noise} gives a quantitative comparison of the deblurring obtained via Algorithm \ref{alg:generative-prior-deblurring} (the free parameters $\lambda$, $\gamma$ and random restarts in the algorithm are fixed as before), and baseline methods CNN, DeblurGAN (trained on fixed 1\% noise level and on varying  1-10\% noise levels) in the presence of Gaussian noise. We also include the performance of deblurred range images $\hat{i}_{\text{range}}$, introduced in Section \ref{sec:Exps-PretrainedPriors}, as a benchmark. Conventional prior based approaches are not included as their performance substantially suffers on noise compared to other approaches. On the vertical axis, we plot the performance metrics (PSNR, and SSIM) and on the horizontal axis, we vary the noise strength from 1 to 10\%. In general, the quality of deblurred range images (expressible by the generators) $\hat{i}_{\text{range}}$ under generative priors surpasses other algorithms on both CelebA, and SVHN. This in a way manifests that under expressive generative priors, the performance of our approach is far superior. The quality of deblurred images $\hat{i}_1$ under generative priors with arbitrary (not necessarily in the range of the generator) input images is the second best on SVHN, however, it under performs on the CelebA dataset; the most convincing explanation of this performance deficit is the range error (not as expressive generator) on the relatively complex/rich images of CelebA. The end-to-end approaches trained on fixed 1\% noise level display a rapid deterioration on other noise levels. Comparatively, the ones trained on 1-10\% noise level, expectedly, show a more graceful performance. DeblurGAN generally under performs compared to our proposed algorithms, however, CNN displays competitive performance, and its deblurred images are second best after $\hat{i}_{\text{range}}$ on CelebA, and third best on SVHN after both $\hat{i}_{\text{range}}$, and $\hat{i}_1$. Qualitative results under heavy noise are depicted in Figure \ref{fig:results-heavynoise}. Our deblurred image $\hat{i}_1$ visually agrees better with $i_{\text{test}}$ than other methods. 

\subsubsection{Random Restarts}
\begin{figure}[t]
\centering
\subfigure[\small CelebA]{\includegraphics[width=0.49\columnwidth]{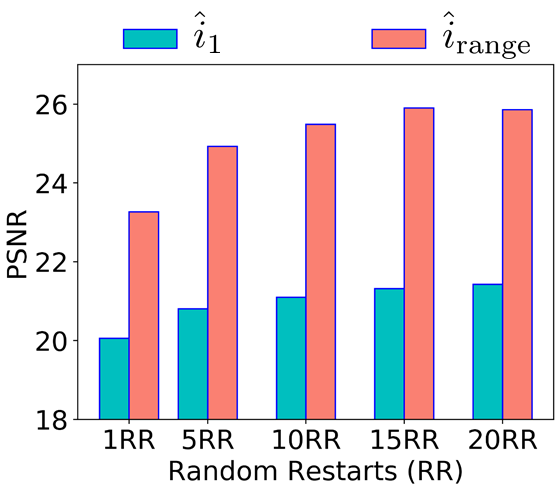}}
\subfigure[\small SVHN]{\includegraphics[width=0.49\columnwidth]{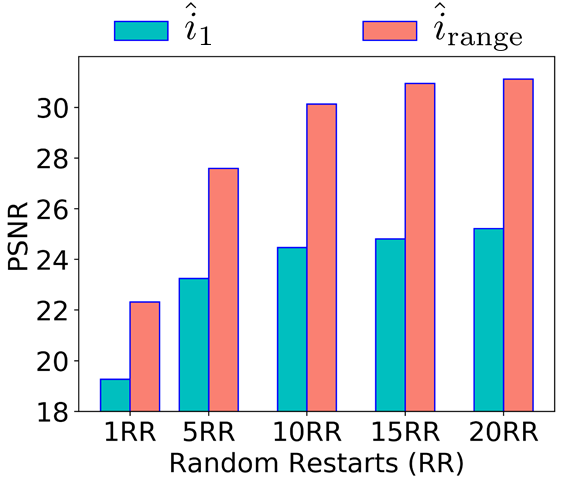}}
\caption{\small Effect of Random Restarts.  Performance of Algorithm \ref{alg:generative-prior-deblurring} for CelebA and SVHN for test images $i_{\text{test}}$ and range images $i_\text{range}$.}
\label{fig:psnr-ssim-randomrestarts}
\end{figure}

Since our proposed algorithms minimize non-convex objectives, the deblurring results depend on the initialization. Higher quality deblurred images are achieved if instead of running the algorithm once, we run it several times each time with a new random initialization of latent dimensions (${z}_i$ and ${z}_k$), and choosing the best  based on the measurement loss (data misfit). Technically, multiple random restarts make us less vulnerable to being trapped in a \textit{not so good} local minima of the non-convex objective by giving the gradient descent algorithm fresh starts. Figure \ref{fig:psnr-ssim-randomrestarts} gives a bar plot of the average PSNR on CelebA and SVHN versus the number of random restarts. Evidently, the PSNR improves with increasing random restarts.

\begin{figure}[h]
\centering
\includegraphics[width=0.6\columnwidth]{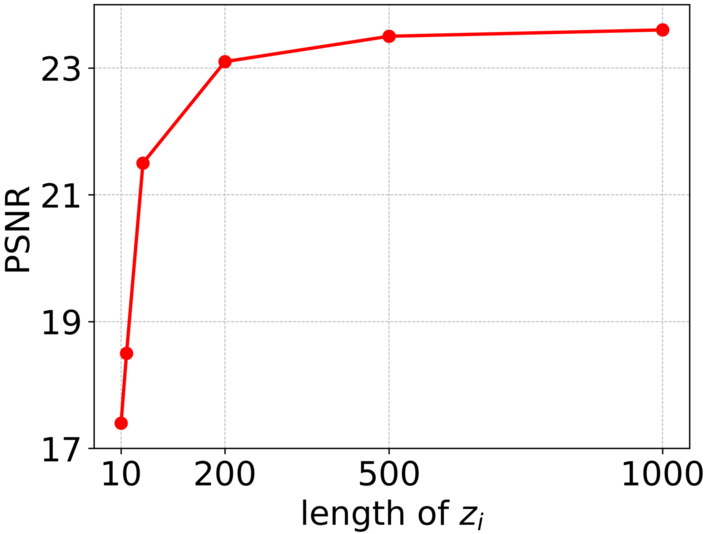}
\caption{Performance of Algorithm \ref{alg:generative-prior-deblurring} with increasing length of $z_i$. A DCGAN was trained on CelebA dataset with varying length of $z_i$. For each case, we plot the average PSNR for Algorithm \ref{alg:generative-prior-deblurring}.}
\label{fig:psnr-zi}
\end{figure}

 \subsubsection{Latent Dimension} The length of the latent parameters also affects the quality of the deblurred image. Figure \ref{fig:psnr-zi}  depicts a relationship between average PSNR of the recovered images $\hat{i}_1$, and the length of $z_i$. We do this by training CelebA image generators with different lengths of $z_i$, and employ each of the trained generator as a prior in Algorithm \ref{alg:generative-prior-deblurring}.  The result shows that increasing the length of $z_i$ above 10 sharply increases the PSNR, which tapers off after the length of $z_i$ exceeds 200. 
 
Increasing the length of parameters $z_i$ gives the generator more freedom to parameterize the latent distribution, and hence better model the underlying random process. Roughly speaking, this results in improving the expressive power of the generator to a certain degree. However, increasing length of $z_i$ also increases the number of unknowns in the deblurring process. Therefore, increasing the length of $z_i$ only improves the performance to a certain degree as depicted in Figure \ref{fig:psnr-zi}. As mentioned in the beginning of experiments that the length of $z_i$ was fixed at 100 in all the performance evaluations above; this plot shows that setting the length of $z_i$ to 200 should roughly improve the average PSNR by 1dB for the deblurred CelebA images.

\subsubsection{Robustness against Large Blurs}
\begin{figure}[t]
\centering
\includegraphics[width=\columnwidth]{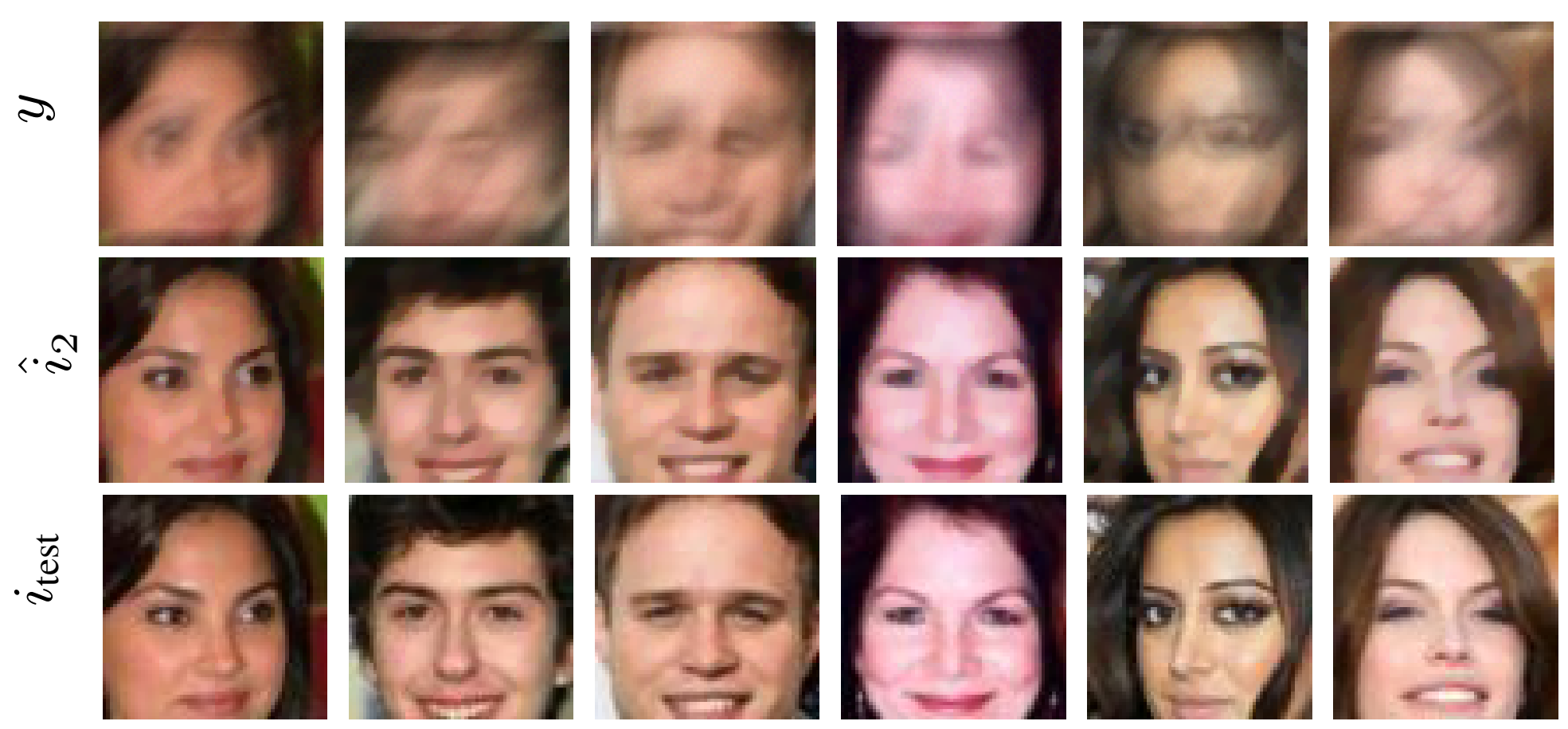}
\caption{\small Large Blurs. Under large blurs, proposed Algorithm \ref{alg:generative+classical-prior-deblurring}, shows excellent deblurring results. }
\label{fig:results-largeblur}
\end{figure}

\begin{figure}[t]
\centering
\includegraphics[width=\columnwidth]{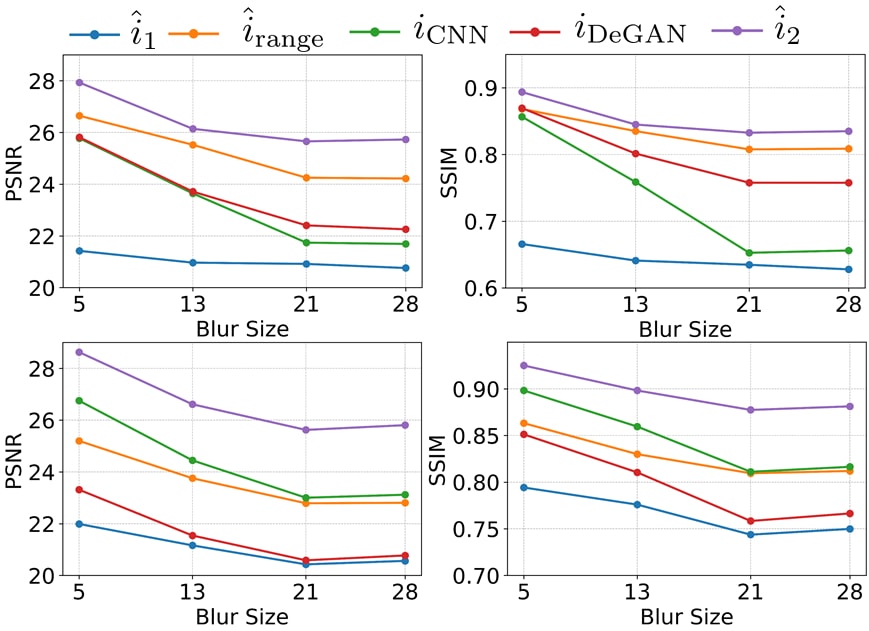}
\caption{\small Blur Size Analysis. Comparative performance of proposed methods, on CelebA (first row) and Shoes (second row), against baseline techniques, as blur length increases.}
\label{fig:psnr-ssim-blursize}
\end{figure}

As is clear from the experiments above that owing to the more involved learning process, the generative priors appear to be far more effective than the classical priors, and firmly guide the deblurring algorithm towards yielding better quality deblurred images. This advantage of generative priors clearly becomes visible in case of large blurs when the blurred image is not even recognizable to the naked eye. Figure \ref{fig:results-largeblur} shows the deblurred images obtained from a very blurry face image. The deblurred image $\hat{i}_2$ using Algorithm 2 above is able to recover the true face from a completely unrecognizable face. The classical baseline algorithms totally succumb to such large blurs. The quantitative comparison against end-to-end neural network based methods CNN, and DeblurGAN is given in Figure \ref{fig:psnr-ssim-blursize}. We plot the blur size against the average PSNR, and SSIM for both Shoes, and CelebA datasets. On both datasets, deblurred images $\hat{i}_2$ using our Algorithm 2 convincingly outperforms all other techniques. For comparison, we also add the performance of  $\hat{i}_{\text{range}}$. Excellent deblurring under large blurs can also be seen in Figure \ref{fig:pg-gan-results} for PG-GAN. To summarize, the end-to-end approaches begin to lag a lot behind our proposed algorithms when the blur size increases. This is owing to the firm control induced by the powerful generative priors on the deblurring process in our newly proposed algorithms. 

\subsection{Extension to Natural Images using Untrained Generators} \label{sec:Exps-UntrainedPriors}

As discussed in detail earlier, the extension of the proposed deblurring under generative priors to more complex/rich natural images is limited by the expressive power of the image generator. Generally, the range error of the image generator deteriorates for relatively more complex/rich image datasets, which in turn results in a below par deblurring performance. One way to address this drawback is to modify Algorithm \ref{alg:generative-prior-deblurring}, which strictly restricts the recovered image to the range of the generator, to Algorithm \ref{alg:generative+classical-prior-deblurring}, which allows some leverage by going beyond the range of the generator under one of the classical priors. However, the question that still remains is that how to extend the deblurring algorithm under generative models alone (without the input from any classical prior as in Algorithm \ref{alg:generative+classical-prior-deblurring})  to complex/rich natural images?


\begin{figure*}[t]
\centering
\includegraphics[width=\textwidth]{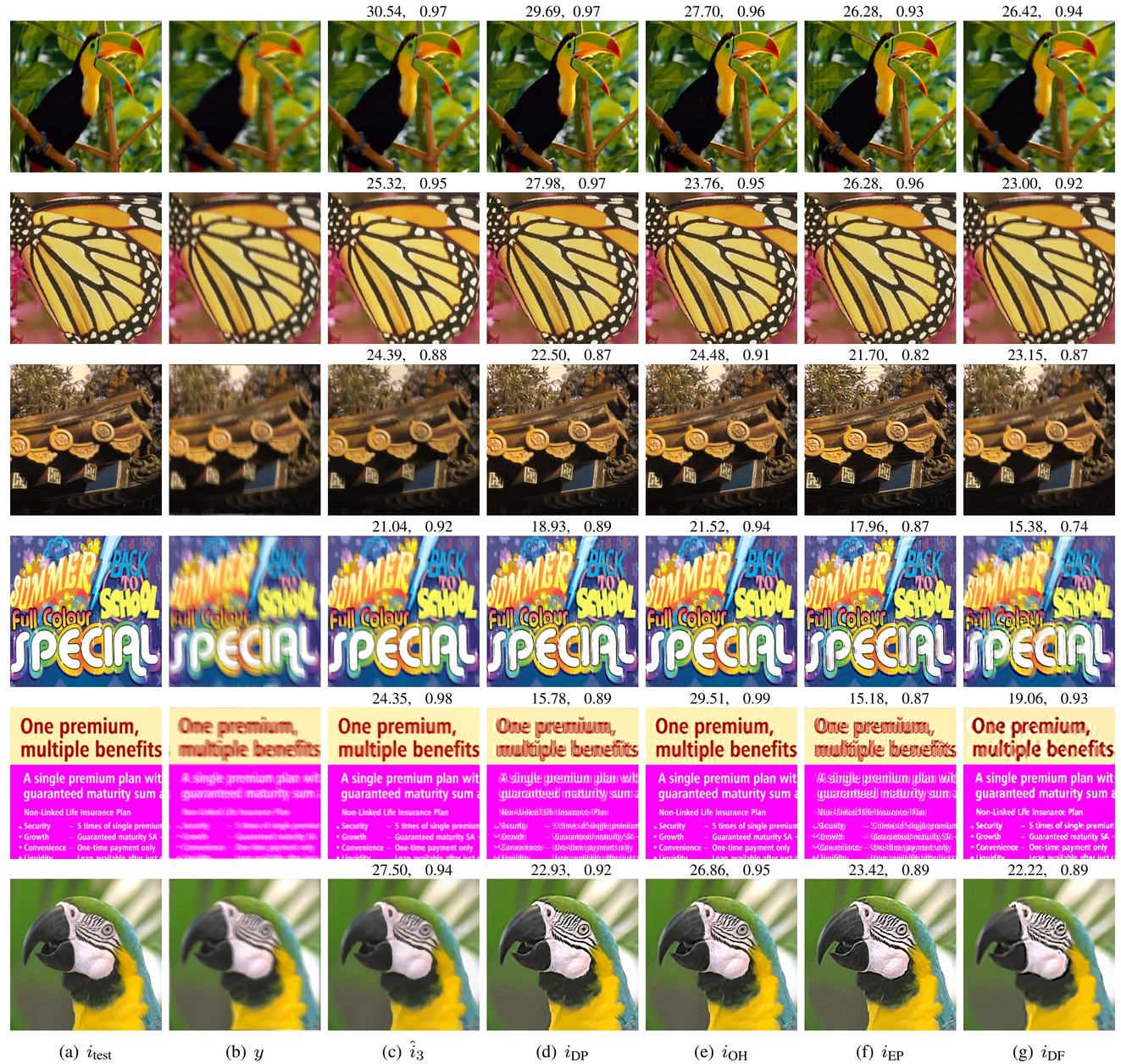}
\caption{ \small Untrained Generative Priors. Deblurring results of arbitrary natural images using an untrained image generator are competitive against the baseline methods. For each deblurred image, PSNR and SSIM are reported at the top.}
\label{fig:DPD-N}
\end{figure*}

To answer the question, one extreme solution to avoid the shortcoming of generative networks on complex images is to completely skip the network training step, and employ untrained generative networks for image as priors. As mentioned, similar ideas has been recently explored in end-to-end networks \cite{ulyanov2017deep}. Algorithm \ref{alg:dpd-n} does exactly this, and updates the weights of a properly initialized network in addition to $z_i$, and $z_k$ in the iterative scheme. We test this algorithm on complex $256 \times 256$ blurred images. A properly initialized DCGAN, see \eqref{eq:deep-prior-weights}, modified to the image resolution, was introduced as an untrained image generator in Algorithm \ref{alg:dpd-n}.{ Initialization of DCGAN was carried out using Adam optimizer with step size set to $0.001$ for $400$ iterations. Later, we optimized the loss in \eqref{eq:dpdn-loss}, again using Adam optimizer for $20,000$ iterations. The step size for updating $z_i$, $z_k$ and $W$ were chosen to be $10^{-3}$, $10^{-3}$ and $10^{-4}$, respectively. Smaller step size for the network weights $W$ is to discourage any large deviation of the weight parameters from our qualified initialization derived from the blurred image; the only available information in this case as the generator is not trained \textit{a priori}. 

Figure \ref{fig:DPD-N} shows the results of Algorithm \ref{alg:dpd-n} on few complex blurry images, and also compares against the classical prior based techniques. Interestingly, even the untrained generator performs quite competitively against these baseline methods. The PSNR, and SSIM values of the deblurred images are also reported in the inset. These initial results are meant to showcase the potential of generative priors on more complex image datasets. This shows that introducing a generative prior in image deblurring is in general a good idea regardless of the expressive power of the generator on the image dataset as it acts as a reasonable prior based on its structure alone. Future work focusing on novel network architecture designs that more strongly favor clear images over blurry ones could pave way for more effective utilization of generative priors in image deconvolution.

\section{Conclusion} \label{sec:conc}

This paper proposes a novel framework for blind image deblurring that uses deep generative networks as priors rather than in a conventional end-to-end manner. We report convincing deblurring results under the generative priors in comparison to the existing methods. A thorough discussion on the possible limitations of this approach on more complex images is presented along with a few effective remedies to address these shortcomings. Importantly, the general strategy of invoking generative priors is not limited to deblurring only but can be employed in other interesting non-linear inverse problems in signal processing, and computer vision. The main contribution of the paper, therefore, goes beyond image deblurring, and is in introducing generative priors as effective method in challenging non-linear inverse problem with a multitude of interesting follow up questions.

\bibliographystyle{IEEEtran}
\bibliography{final}


\newpage
\onecolumn
\section*{Appendix}

Extended qualitative results for SVHN, CelebA and Shoes are provided in Figures \ref{fig:svhn-results-appendix}, \ref{fig:celebA-results-appendix} and \ref{fig:shoes-results-appendix}, respectively. In addition to motion blurs, we also trained a generative model $G_\setK$ on Gaussian blurs and show qualitative results for Algorithm 1 on SVHN and CelebA in Figure \ref{fig:svhn-gaussian-appendix} and \ref{fig:celeba-gaussian-appendix}, respectively. Results of PG-GAN are also made available in Figure \ref{fig:pggan-results-appendix}.


\begin{figure*}[h]
\centering
\includegraphics[width=\textwidth]{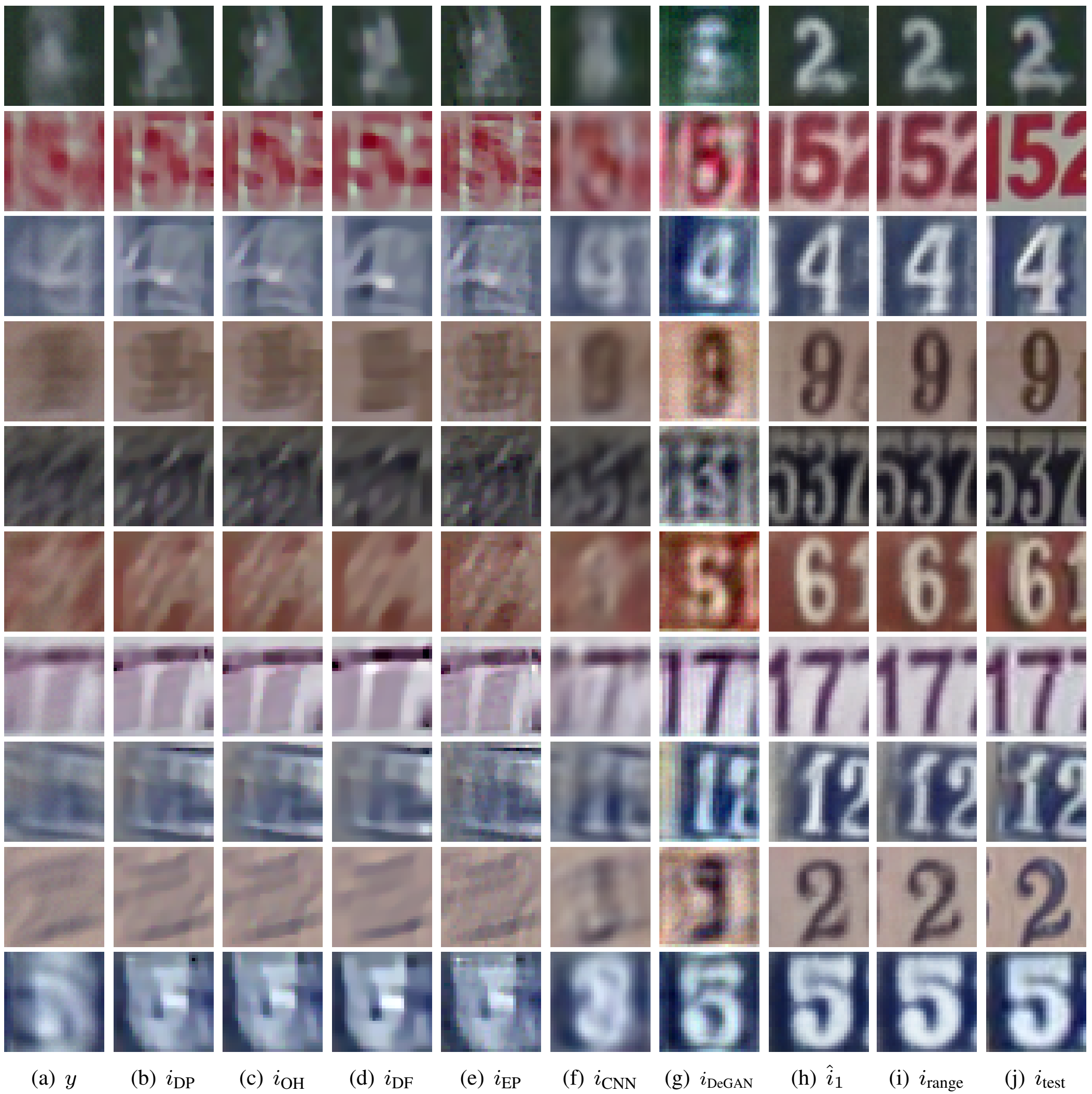}
\caption{\small{Comparison of image deblurring on SVHN for Algorithm 1 with baseline methods. Deblurring results of Algorithm 1, $\hat{i}_1$, are  superior than all other baseline methods, especially under large blurs. Better results of Algorithm 1 on SVHN are explained by the close proximity between range images $i_\text{range}$ and original groundtruth images $i_\text{test}$.}}
\label{fig:svhn-results-appendix}
\end{figure*}

\begin{figure*}[h]
\centering
\includegraphics[width=\columnwidth]{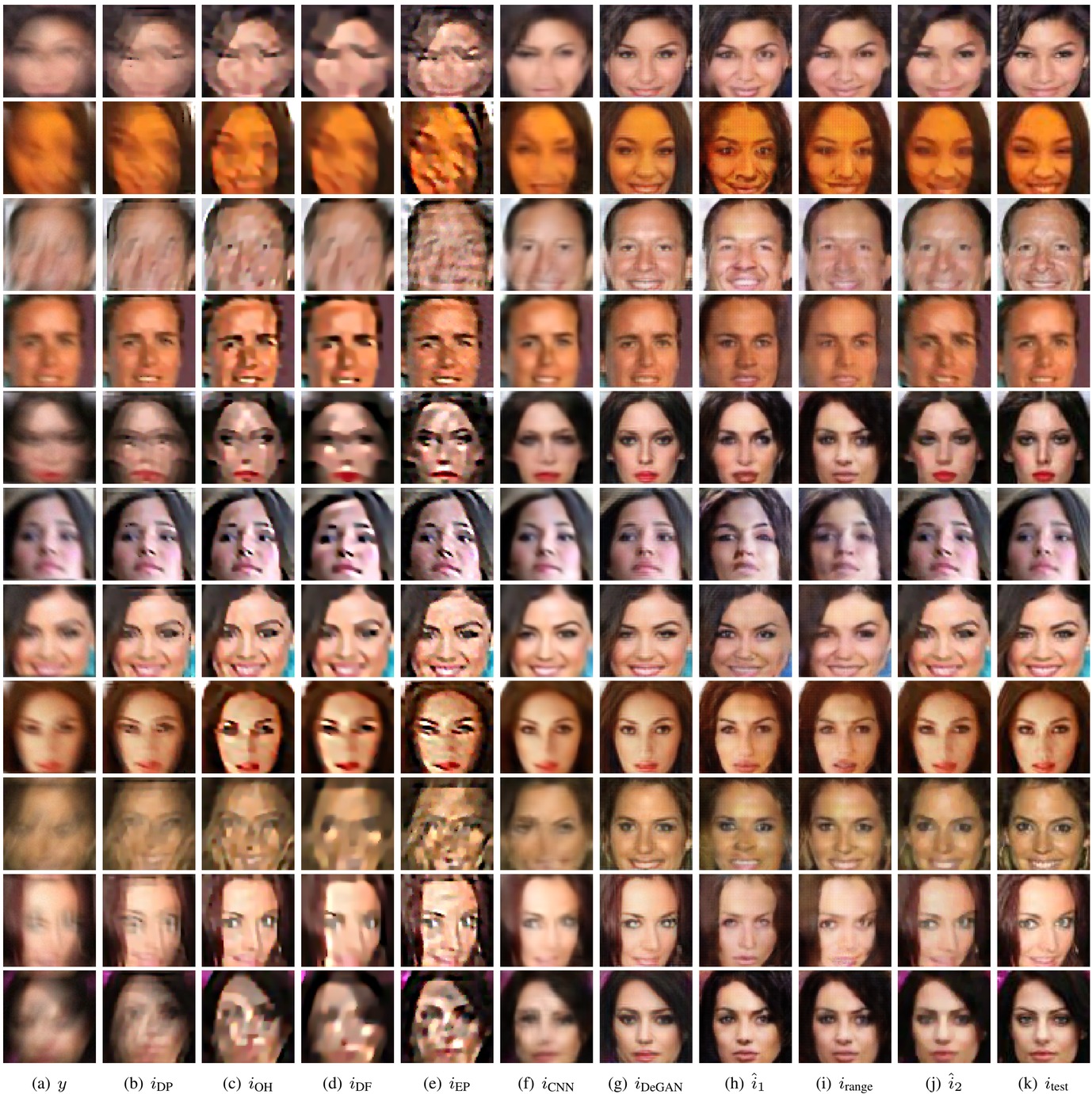}
\caption{\small{Comparison of image deblurring on CelebA for Algorithm 1 and 2 with baseline methods. Deblurring results of Algorithm 2, $\hat{i}_2$, are superior than all other baseline methods, especially under large blurs. Deblurred images of DeblurGAN, $i_{\text{DeGAN}}$, although sharp, deviate from the original images, $i_\text{test}$, whereas Algorithm 2 tends to agree better with the groundtruth.}}
\label{fig:celebA-results-appendix}
\end{figure*}

\begin{figure*}[p]
\centering
\includegraphics[width=\columnwidth]{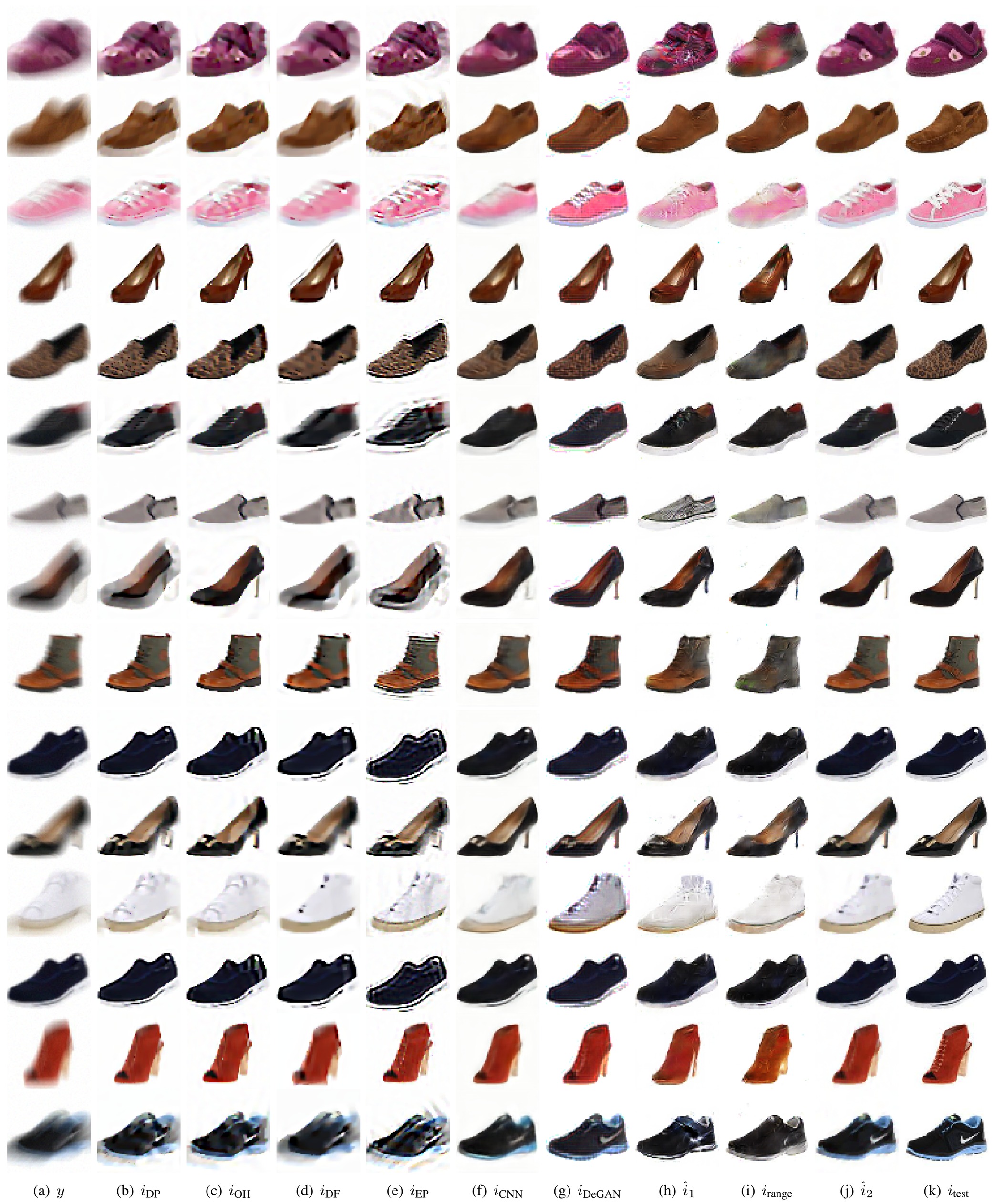}
\caption{\small{Comparison of image deblurring on Shoes for Algorithm 1 and 2 with baseline methods. Deblurring results of Algorithm 2, $\hat{i}_2$, are superior than all other baseline methods, especially under large blurs. Deblurred images of DeblurGAN, $i_{\text{DeGAN}}$, although sharp, deviate from the original images, $i_\text{test}$, whereas Algorithm 2 tends to agree better with the groundtruth.}}
\label{fig:shoes-results-appendix}
\end{figure*}

\clearpage
\begin{figure*}
\centering
\includegraphics[width=0.45\textwidth]{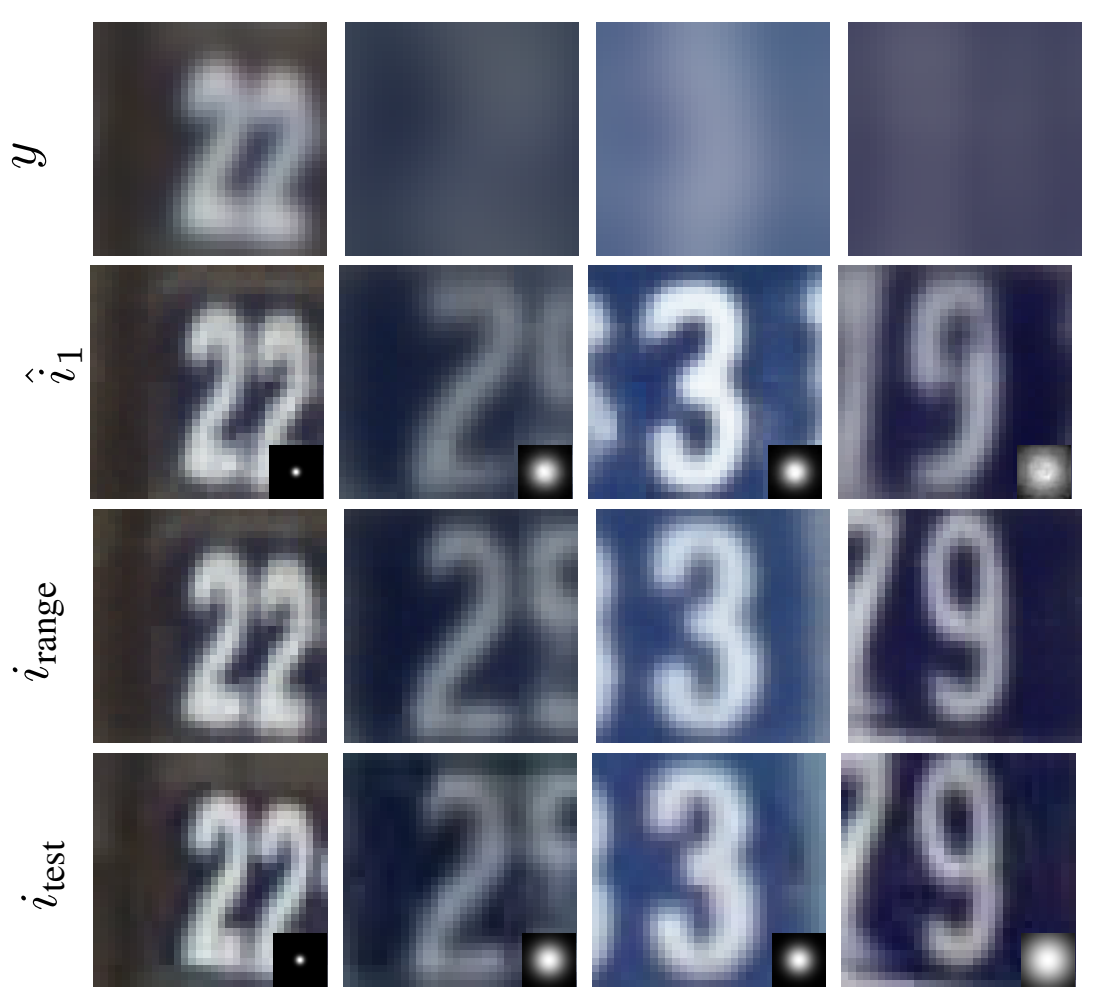}
\caption{\small{Image deblurring results for Algorithm 1 on SVHN with Gaussian blurs. Images from test set along with corresponding blur kernels are convolved to produce blurry images.}}
\label{fig:svhn-gaussian-appendix}
\end{figure*}

\begin{figure*}[h]
\centering
\includegraphics[width=0.45\textwidth]{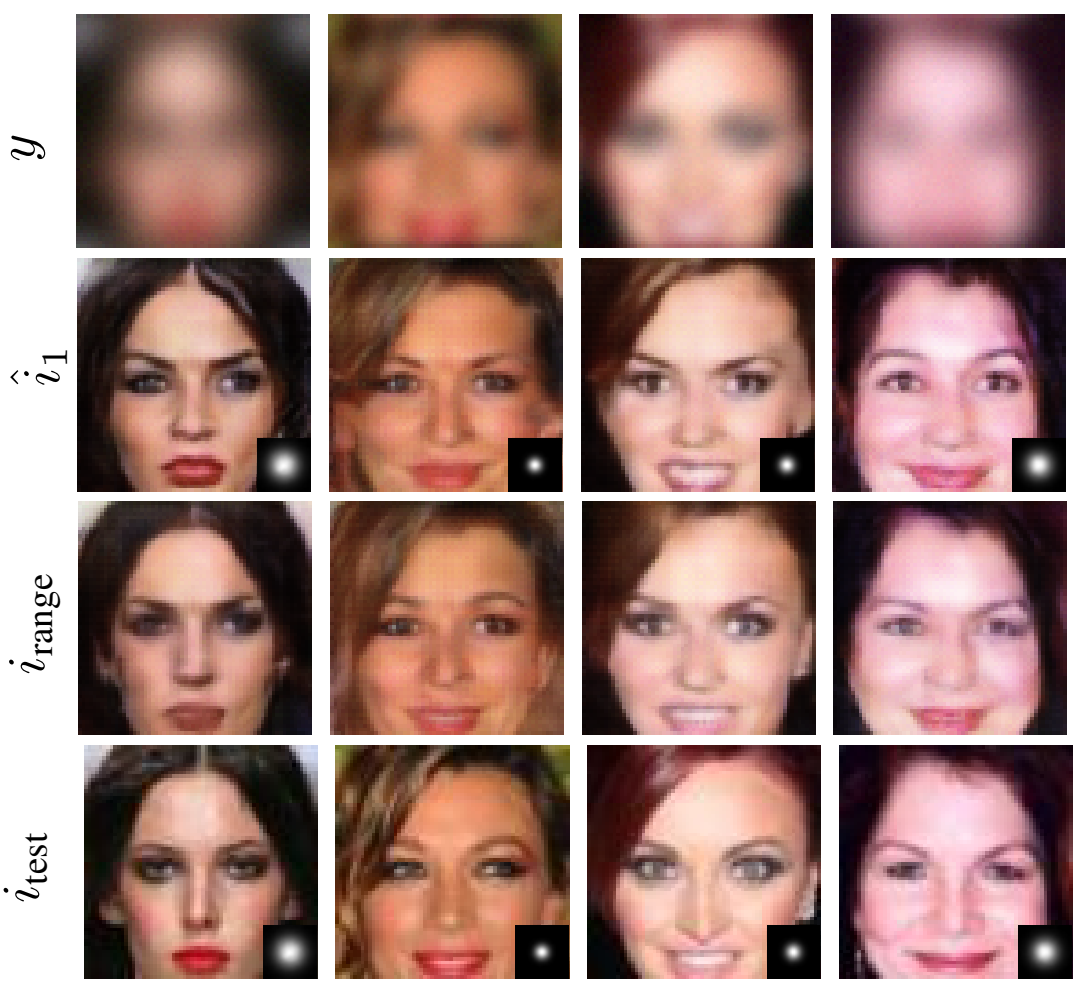}
\caption{\small{Image deblurring results using Algorithm 1 on CelebA for Gaussian blurs. Images from test set along with corresponding blur kernels are convolved to produce blurry images.}}
\label{fig:celeba-gaussian-appendix}
\end{figure*}

\begin{figure*}[h]
\centering
\includegraphics[width=0.6\textwidth]{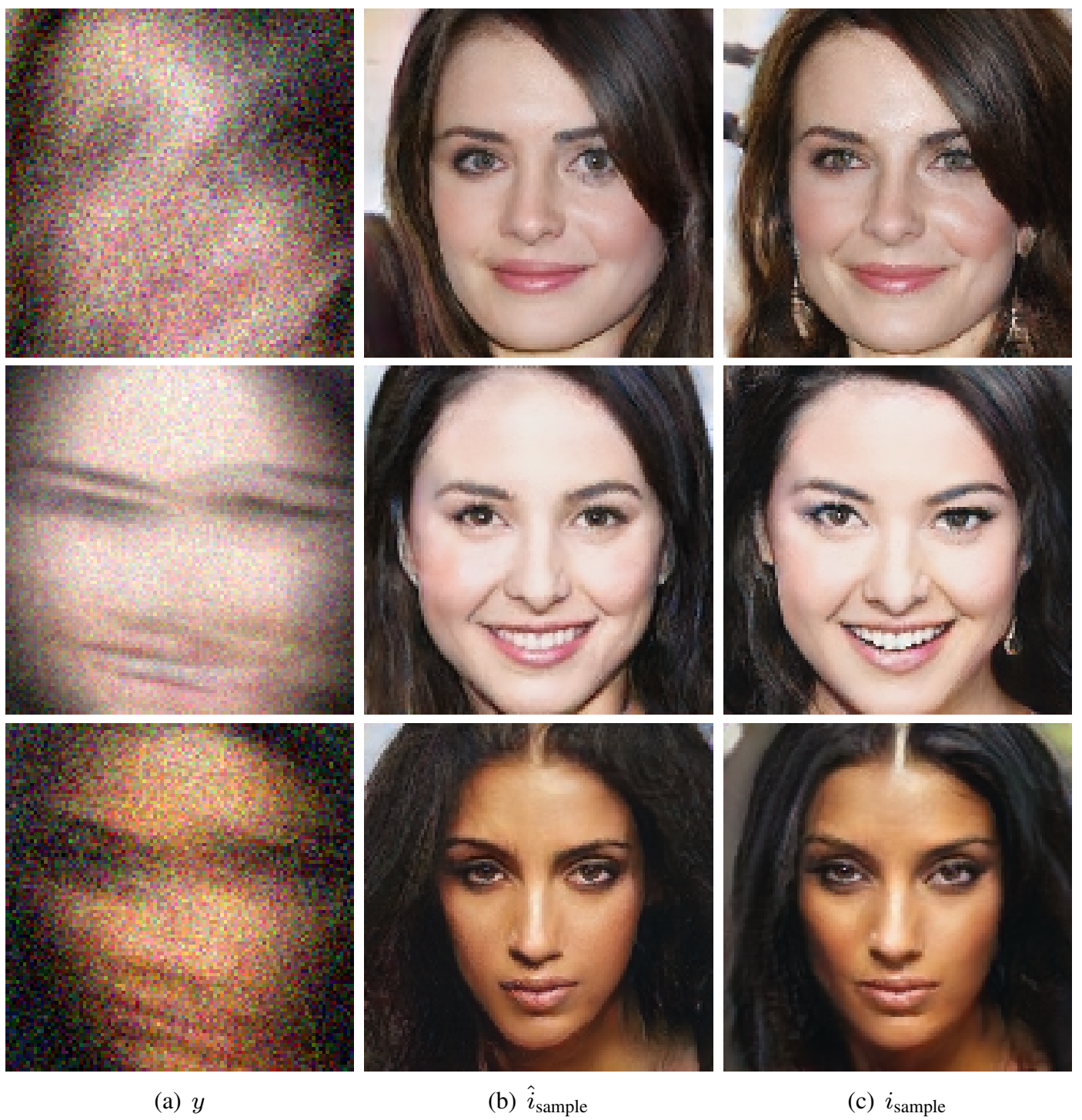}
\caption{\small{Image deblurring results using PG-GAN as generator $G_\setI$ under heavy noise. Images sampled from PG-GAN, were blurred, and Algorithm 1 was used to deblur these blurry images.}}
\label{fig:pggan-results-appendix}
\end{figure*}

\end{document}

%% file: Ahmed-defs.tex
\newcommand{\R}{\mathbb{R}}

\newcommand{\relu}{\operatorname{relu}}
\newcommand{\E}{\operatorname{E}}

\newcommand{\set}[1]{\mathcal{#1}}

\newcommand{\setI}{\set{I}}

\newcommand{\setK}{\set{K}}
\newcommand{\setL}{\set{L}}

\newcommand{\setN}{\set{N}}